\documentclass{article}

\usepackage{iclr2027_conference,times}

\usepackage{amsmath,amsfonts,bm}

\def\eqref#1{equation~\ref{#1}}

\def\1{\bm{1}}

\def\vmu{{\bm{\mu}}}
\def\vtheta{{\bm{\theta}}}

\def\vh{{\bm{h}}}

\def\vp{{\bm{p}}}
\def\vq{{\bm{q}}}

\def\vs{{\bm{s}}}

\def\vx{{\bm{x}}}

\def\vz{{\bm{z}}}

\def\mW{{\bm{W}}}

\DeclareMathAlphabet{\mathsfit}{\encodingdefault}{\sfdefault}{m}{sl}
\SetMathAlphabet{\mathsfit}{bold}{\encodingdefault}{\sfdefault}{bx}{n}

\usepackage{graphicx}
\usepackage{subfig}
\usepackage{amssymb,mathrsfs,amsmath,pifont}
\usepackage{xcolor}
\usepackage[colorlinks=true, citecolor=blue]{hyperref}
\usepackage{booktabs,multirow,graphicx,xcolor,colortbl}%
\usepackage{needspace}
\usepackage{amsthm}%

\usepackage{nicematrix}%

\usepackage{url}

\usepackage{algorithm}%
\usepackage{algpseudocode}%

\title{Do Not Train Away Uncertainty: Early Uncertainty Anchored Calibration}

\author{\parbox[t]{\dimexpr\textwidth-2\tabcolsep\relax}{\raggedright
\mbox{\textbf{Yutong Xie}\textsuperscript{\bfseries 1}},
\mbox{\textbf{Jiawei Tang}\textsuperscript{\bfseries 2}},
\mbox{\textbf{Zhenglin Hua}\textsuperscript{\bfseries 2}},
\mbox{\textbf{Yuxiang Ma}\textsuperscript{\bfseries 1}},
\mbox{\textbf{Si Qin}\textsuperscript{\bfseries 2}},
\mbox{\textbf{Yaxin Hou}\textsuperscript{\bfseries 2}},
\mbox{\textbf{Hui Liu}\textsuperscript{\bfseries 4}},
\mbox{\textbf{Junhui Hou}\textsuperscript{\bfseries 5}},
\mbox{\textbf{Yuheng Jia}\textsuperscript{\bfseries 2,3,4}{\boldmath\thanks{Corresponding author.}}}
} \\
\textsuperscript{1}School of Software Engineering, Southeast University, Nanjing, China \\
\textsuperscript{2}School of Computer Science and Engineering, Southeast University, Nanjing, China \\
\textsuperscript{3}Key Laboratory of New Generation Artificial Intelligence Technology and Its \\
Interdisciplinary Applications (Southeast University), Ministry of Education, China \\
\textsuperscript{4}School of Computing Information Sciences, Saint Francis University, Hong Kong, China \\
\textsuperscript{5}Department of Computer Science, City University of Hong Kong, Hong Kong, China \\
\texttt{\{xyt,jw\_tang,220252358,220256453,siqin,yaxin,yhjia\}@seu.edu.cn} \\
\texttt{h2liu@sfu.edu.hk, jh.hou@cityu.edu.hk}
}

\iclrfinalcopy
\begin{document}

\maketitle
\lhead{Preprint}

\begin{abstract}
Deep neural networks, including large language models, have achieved remarkable performance across various tasks. However, they are prone to overconfidence during training or fine-tuning. In this work, we observe a consistent phenomenon across different models that the early model is better calibrated, while later training or fine-tuning yields marginal accuracy gains but substantially increases calibration errors. Our analysis suggests that the early model retains uncertainty awareness in both its predictions and features, which is gradually lost with continued training. To avoid training away this uncertainty awareness, we propose \textbf{EUA-Cal}, a novel method that exploits the \textbf{E}arly model as an \textbf{U}ncertainty \textbf{A}nchor for \textbf{Cal}ibration. EUA-Cal introduces early prediction regularization to preserve early predictive uncertainty and prototype structure regularization to exploit uncertainty reflected in the early feature space, jointly mitigating overconfidence. Extensive experiments on image classification and multiple-choice question answering across eight diverse models demonstrate that EUA-Cal outperforms state-of-the-art calibration methods.

\end{abstract}

\section{Introduction}
\label{intro}
In recent years, deep neural networks (DNNs), including large language models (LLMs), have achieved significant progress in image classification \citep{wrn}, object segmentation \citep{sam}, and natural language processing \citep{transformer}. Despite their remarkable performance, these models are prone to overconfidence during training or fine-tuning, with the predicted confidence exceeding the actual accuracy \citep{ts,cllm}. This overconfidence reflects an underestimate of the uncertainty in their predictions and can reduce their reliability, particularly in high-risk applications such as autonomous driving \citep{Zhang_2025_ICCV} and medical diagnosis \citep{10.1609/aaai.v39i4.32396}.

To calibrate model confidence, various calibration techniques have been proposed, including post-hoc calibration \citep{ts,NeuralCalibration}, loss function and regularization designs \citep{ls,focal,mbls,balcal}, mixup-based methods \citep{mixup,skmixup}, uncertainty-aware methods \citep{edl,ibedl}, etc. Expected Calibration Error (ECE) \citep{ts} is a commonly used metric for evaluating the calibration performance of these methods, which measures the discrepancy between predicted confidence and empirical accuracy across confidence bins. A lower value of ECE indicates better calibration performance, i.e., the model is aware of how uncertain its predictions are.

\begin{figure}[t]
    \centering
    \subfloat[]
    {
        \includegraphics[width=0.32\textwidth]{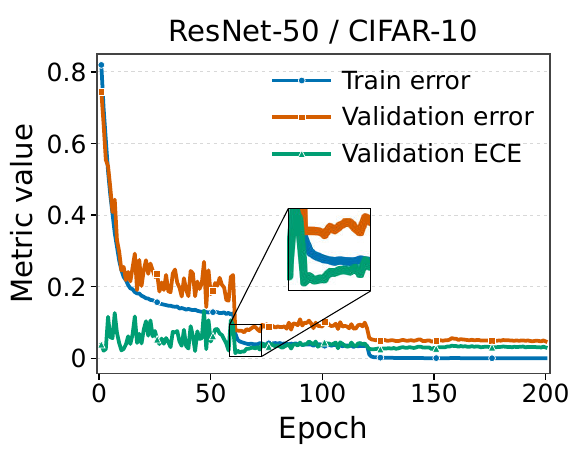}
        \includegraphics[width=0.32\textwidth]{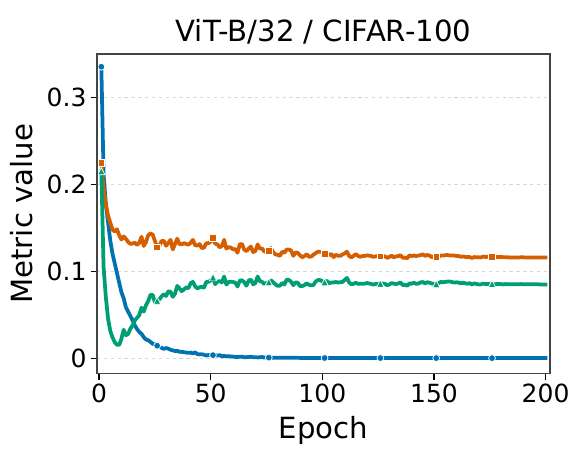}
        \includegraphics[width=0.32\textwidth]{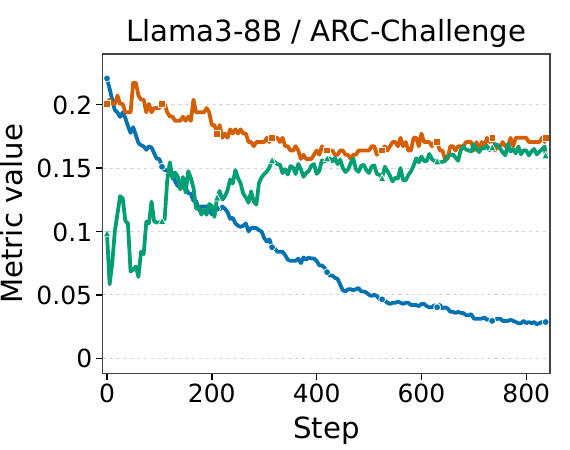}
        \label{1a}
    }\par
    \centering
    \subfloat[]{%
        \hspace{4.3mm}
        \includegraphics[width=0.28\linewidth]{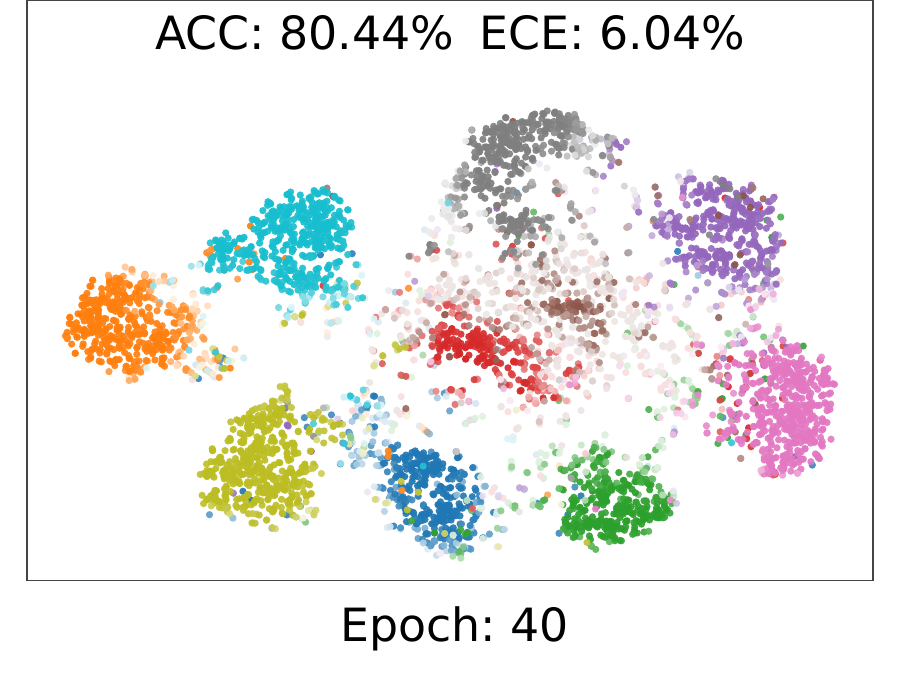}%
        \hspace{5.8mm}
        \includegraphics[width=0.28\linewidth]{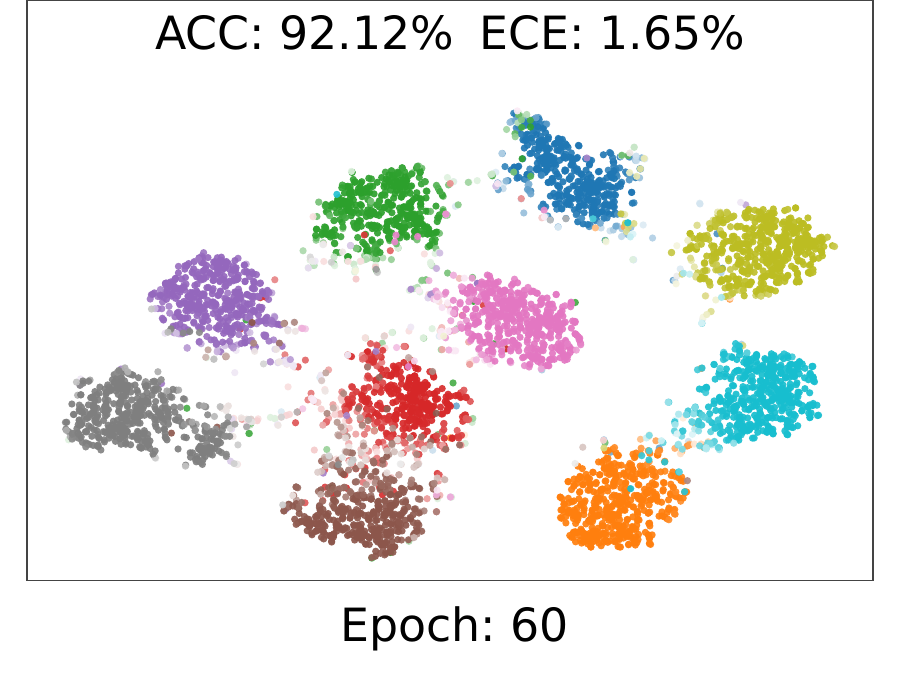}%
        \hspace{5.8mm}
        \includegraphics[width=0.28\linewidth]{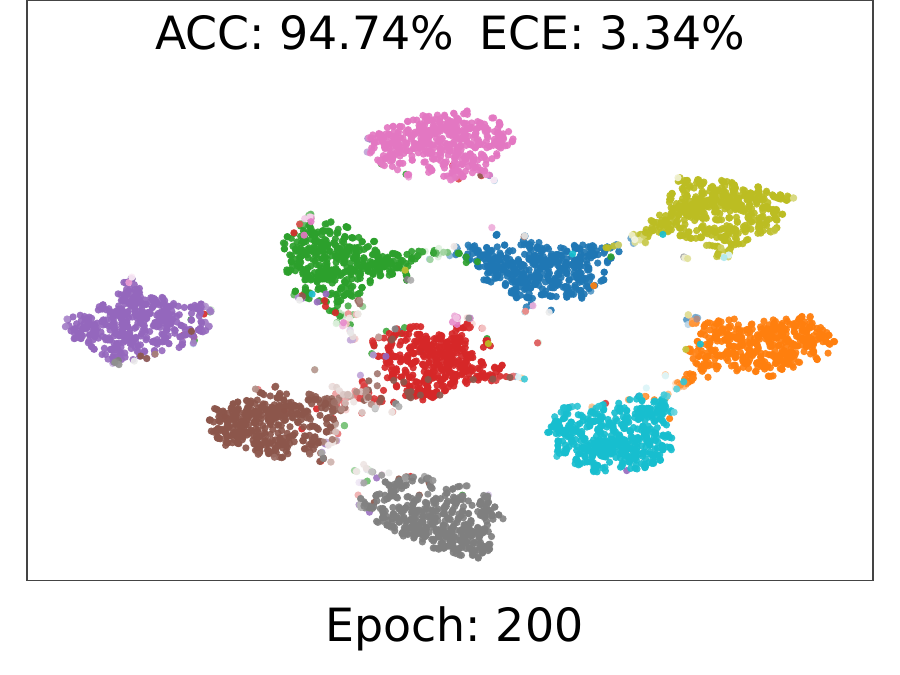}%
        \label{1b}
    }

    \caption{\textbf{(a)} Training and validation classification errors and validation ECE for ResNet-50 on CIFAR-10, ViT-B/32 on CIFAR-100, and Llama3-8B on ARC-Challenge. \textbf{(b)}: t-SNE visualizations of features from ResNet-50 on CIFAR-10 at epochs 40, 60, and 200. Lighter shades indicate features associated with lower confidence. More examples are provided in Appendix~\ref{vis}.} 
    \label{fig1}
\end{figure}

Some studies \citep{ts,focal} have shown that DNNs tend to produce overconfident predictions with continued training. We observe a similar pattern across various models, where the early model is better calibrated but its calibration worsens as training continues. Figure~\ref{fig1}(a) illustrates the classification errors of three different models on the train and validation sets, along with the ECE on the validation set. On all three models, we observe that ECE reaches its minimum at an early stage of training or fine-tuning, but then increases in later stages, while accuracy improves only marginally. This indicates that there exists an early model that achieves the lowest ECE and exhibits the greatest uncertainty awareness in its predictions. However, this awareness is gradually lost during subsequent training, resulting in worse calibration performance.

Beyond predictions of the early model, we observe that it also exhibits uncertainty awareness in feature structure. Figure~\ref{fig1}(b) visualizes the feature structure and the corresponding calibration performance of ResNet-50 at different training epochs. We find that features with low confidence (shown in lighter shades) are crucial for calibration. At epoch 40, many features have low confidence, but their cluster structure remains poorly defined. The model exhibits the lowest accuracy and the highest ECE. At epoch 60, corresponding to the early model in Figure~\ref{fig1}(a), clearer clusters emerge while features with low confidence remain near their boundaries. At this stage, accuracy improves and ECE reaches its minimum. By epoch 200, the model achieves the highest accuracy with the most compact clusters. However, features with low confidence almost disappear, while the ECE increases. These observations suggest that the early model with the best ECE forms a clear feature structure while retaining some low confidence features, demonstrating the best uncertainty awareness in its features. However, this uncertainty awareness also fades as training continues.%

 To avoid training away this uncertainty awareness in predictions and features, we propose \textbf{EUA-Cal}, a novel method that exploits the \textbf{E}arly model as an \textbf{U}ncertainty \textbf{A}nchor for \textbf{Cal}ibration. Specifically, in the pre-learning stage, we obtain the early model by stopping training at a specific epoch, which is selected on validation sets. We then introduce \emph{Early Prediction Regularization} (EPR) to preserve early predictive uncertainty and \emph{Prototype Structure Regularization} (PSR) to exploit uncertainty awareness in features of the early model. Finally, we combine EPR and PSR to formulate the overall EUA-Cal objective, aiming to preserve uncertainty and mitigate overconfidence.

Our contributions are summarized as follows:
\begin{itemize}
\item We identify a general phenomenon in training or fine-tuning where the early model is better calibrated. Our analysis further shows that the early model preserves uncertainty awareness in both the predictions and the feature structure. However, this uncertainty is gradually lost as training continues, resulting in overconfidence.
\item We propose EUA-Cal, a novel method that exploits the uncertainty awareness retained by the early model as an anchor. Our method is achieved through regularization terms that EPR aligns current predictions with early predictions and PSR derives probability distributions from early feature structures, jointly mitigating overconfidence.
\item Our method achieves superior performance in calibration across diverse network architectures, including deep neural networks such as ResNet and DenseNet, as well as Transformer-based models such as ViT, Llama, and Qwen. Extensive experiments on eight models and six datasets demonstrate that EUA-Cal outperforms state-of-the-art methods. 

\end{itemize}

\section{Related Work}
\label{Related Work}
\textbf{Calibration for DNNs.} Calibration aims to ensure that predicted confidence is indicative of the actual likelihood of correctness \citep{survey}. Post-hoc methods learn a calibration function on a validation set \citep{beta,ts,NeuralCalibration}. Among them, Temperature Scaling \citep{ts} is a widely used method that learns a single temperature to adjust the output logits. Despite their simplicity and effectiveness, post-hoc methods perform poorly under distribution shifts \citep{canyou}. Loss function and regularization designs reduce overconfidence by implicitly or explicitly maximizing the entropy of predictive distributions. Label Smoothing (LS) \citep{ls} softens one-hot labels, while Focal Loss (FL) \citep{focal} implicitly increases the entropy of the predicted distribution. MbLS \citep{mbls} controls the margin on logit distances to prevent the model from non-informative predictions. BalCAL \citep{balcal} combines a fixed ETF classifier with a standard linear classifier to adjust confidence. Mixup-based methods were originally proposed to improve generalization and have been proven effective for calibration \citep{mixup,rankmixup,skmixup}. \citet{mit-a} note that the confidence penalty of Mixup would harm calibration and propose MIT to avoid the negative impact of the confidence penalty. Several methods propose calibration methods for specific network structures \citep{tst,plp,maccal}, but their generalizability across different structures remains unclear.

\textbf{Calibration for Fine-tuned LLMs.} Recent studies have focused on fine-tuned LLM calibration \citep{cllm,bayesian}. Traditional methods, such as MC-Dropout \citep{mc-dropout} and Deep Ensemble \citep{des} provide uncertainty information but require multiple inferences when applied to fine-tuned LLMs. Several methods originally developed for the calibration of DNNs, such as Mixup and LS, have also been shown to effectively calibrate LLMs during fine-tuning \citep{mixup_plm,ls_llm}. Compared with these methods, our method is also applicable to fine-tuned LLMs and achieves superior performance. Distillation \citep{distillation} improves calibration by knowledge distillation from calibrated model ensembles. However, this approach relies on an ensemble of multiple models and an additional validation set for calibration. IB-EDL \citep{ibedl} introduces an information bottleneck into the EDL \citep{edl} framework to suppress spurious and excessive evidence, thus mitigating overconfidence in fine-tuned LLMs.

\section{Methodology}
\label{Methodology}
\begin{figure}[t]
    \centering
        \includegraphics[width=0.98\linewidth]{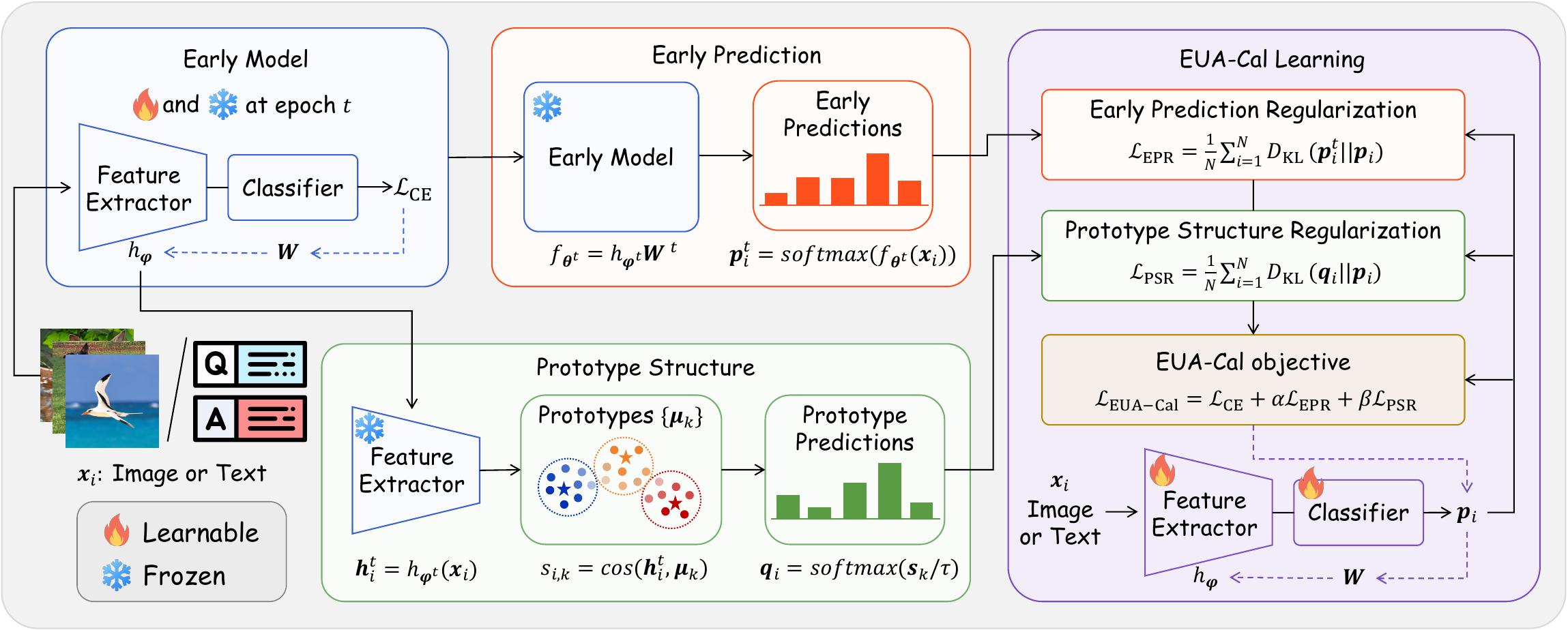}
    \caption{An illustration of the proposed EUA-Cal approach for calibration. In the pre-learning stage, we freeze the early model as an uncertainty anchor at the selected training epoch. Its predictions are used for EPR, while confidence-weighted class prototypes constructed from early features provide structural targets for PSR. The overall objective guides training or fine-tuning to preserve uncertainty and mitigate overconfidence.}
    \label{fig2}
\end{figure}

In this section, we first provide the problem statement of model calibration in Sec. \ref{3.1} and then introduce the proposed method, regarding details of early prediction regularization, prototype structure regularization, and the overall EUA-Cal objective in Sec. \ref{EUA-Cal}. The overview of our method is shown in Figure \ref{fig2}.

\subsection{Problem Statement}
\label{3.1}
We consider a $K$-class classification problem with a training set $\mathcal{D}=\{(\vx_i,y_i)\}_{i=1}^N$, where $\vx_i\in\mathcal{X}$ is a training sample, $y_i\in\mathcal{Y}$ is its corresponding label and $N$ denotes the total number of samples. Let $f_{\vtheta}$ be a model parameterized by $\vtheta$ that produces a logit $\vz_i=f_{\vtheta}(\vx_i)\in\mathbb{R}^K$, and the class probabilities are defined as $\vp_i=\{p_{i,k}\}_{k=1}^K=softmax(\vz_i)$, where $p_{i,k}$ is the predicted probability for class $k$. Based on these probabilities, the predicted label and confidence are defined as $\hat{y}_i=\arg \max_k p_{i,k}$ and $\hat{p}_i=\max_k p_{i,k}$, respectively. We further separate the model $f_{\vtheta}$ into a feature extractor $h_{\boldsymbol{\varphi}}:\mathcal{X}\rightarrow\mathbb{R}^d$ and a linear classifier $\mW\in\mathbb{R}^{d\times K}$, such that $f_{\vtheta}(\vx_i)=h_{\boldsymbol{\varphi}}(\vx_i)\mW$. During training, we denote the parameters obtained at the $t$-th epoch by $\vtheta^{t}=\{\boldsymbol{\varphi}^{t}, \mW^{t}\}$, and the corresponding model is $f_{\vtheta^{t}}(\cdot)=h_{\boldsymbol{\varphi}^{t}}(\cdot)\mW^{t}$. 

A model is well-calibrated if the predicted confidence $\hat{p}_i$ reflects the actual probability of correctness, i.e., $\mathbb{P}(\hat{y}_i=y_i|\hat{p}_i=p)=p$ for all $p\in[0,1]$. However, DNNs trained or LLMs fine-tuned with CE loss tend to be overconfident, producing higher confidence than its accuracy. Confidence calibration aims to align predicted confidence with actual accuracy, thereby producing more reliable predictions.  

\subsection{Early Uncertainty Anchored Calibration}
\label{EUA-Cal}
Based on previous analysis, we propose EUA-Cal, a novel method that exploits uncertainty awareness retained by the early model for calibration.

\textbf{Early Prediction Regularization.} As analyzed in Sec. \ref{intro}, prediction of the early model is better calibrated and retains reliable uncertainty awareness for samples. Motivated by this insight, we introduce \emph{Early Prediction Regularization} (EPR) to preserve the specific uncertainty of samples captured at the early training stage. Formally, given the parameters $\vtheta^{t}$ from the selected training iteration $t$, we freeze the corresponding model $f_{\vtheta^{t}}$. The EPR loss is defined as the Kullback-Leibler (KL) divergence between the early and current prediction distributions, i.e.,
\begin{equation}
    \mathcal{L}_{\mathrm{EPR}}=\frac{1}{N}\sum_{i=1}^ND_{\mathrm{KL}}(\vp^t_i||\vp_i)=\frac{1}{N}\sum_{i=1}^N\sum_{k=1}^Kp^t_{i,k}\log \left( \frac{p^t_{i,k}}{p_{i,k}} \right),
    \label{eq1}
\end{equation}
where $\vp^t_i=\{p_{i,k}^t\}_{k=1}^K=\text{softmax}(f_{\vtheta^{t}}(\vx_i))$ is the early prediction distribution of sample $\vx_i$ from the frozen early model. The traditional CE loss with hard label continuously pushes all samples toward one-hot prediction, regardless of the sample specific uncertainty. In contrast, early predictions retain sample uncertainty before the overconfidence emerges. EPR utilizes them as reliable references to constrain excessive confidence growth on uncertain samples, while preserving confident predictions for certain samples. 

\begin{figure}[t]
    \centering

    \subfloat[]{%
        \includegraphics[width=0.35\linewidth]{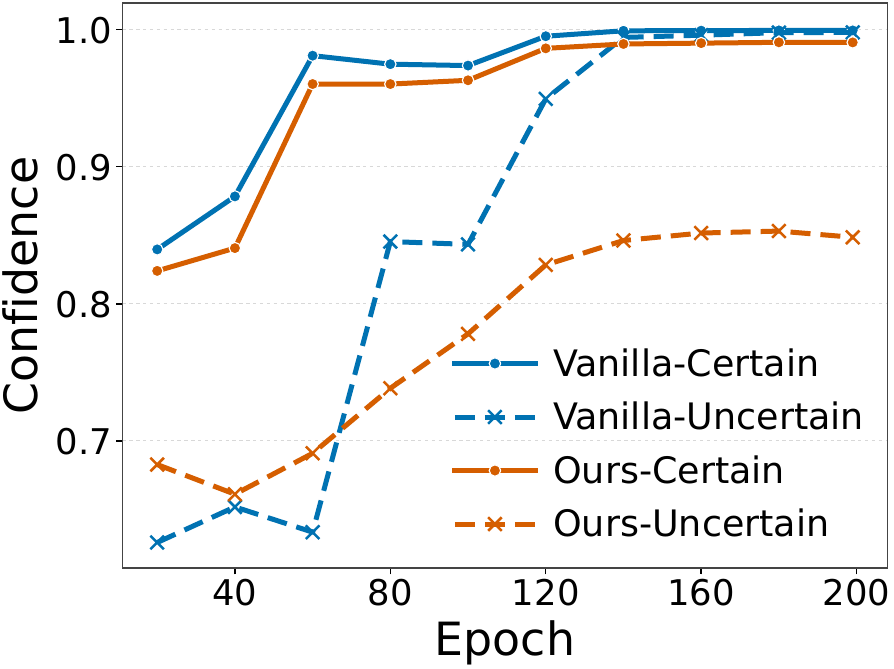}
        \label{2a}
    }\hspace{8mm}
    \subfloat[]{%
        \includegraphics[width=0.42\linewidth]{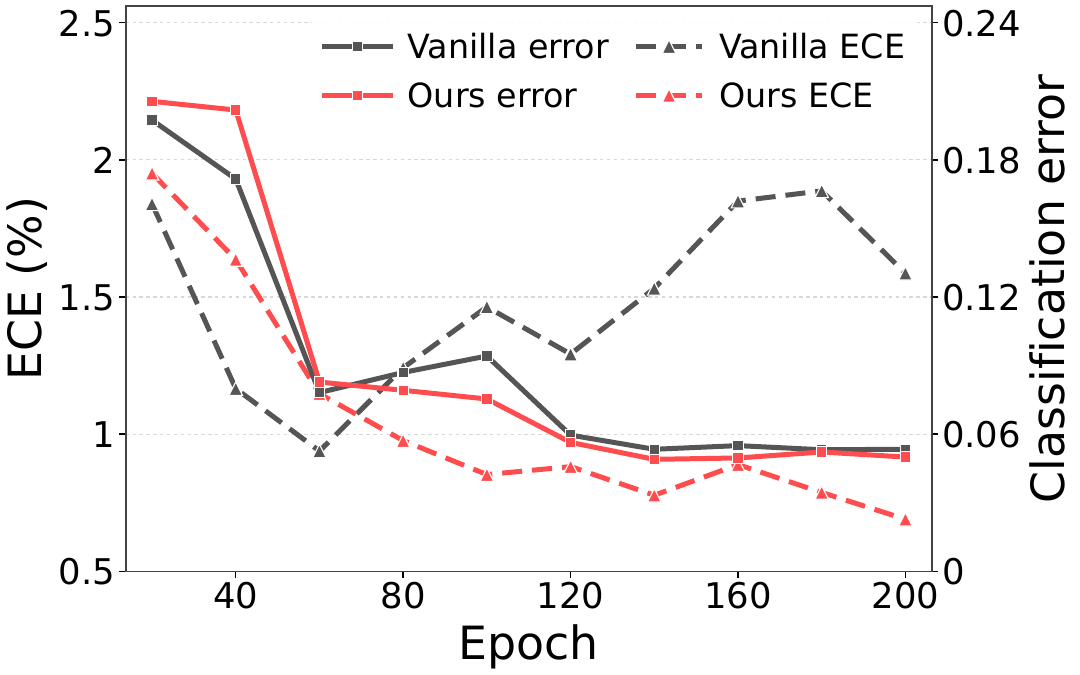}%
        \label{2b}
    }

    \caption{\textbf{(a)} Confidence of certain and uncertain samples under the vanilla CE model and our method. Samples are divided according to the confidence of the selected early model. \textbf{(b)} Classification errors and ECE of the prototype predictions on the validation set using prototypes constructed from different training epochs.}
    \label{fig3}
\end{figure}

To verify the role of uncertainty awareness in early predictions, we divide the samples into certain and uncertain groups according to the confidence of the selected early model (e.g., epoch 60 on ResNet-50). Figure \ref{fig3}(a) illustrates the average confidence of CIFAR-10 samples. For certain samples, confidence remains high throughout training. However, samples identified as uncertain at the early stage gradually become highly confident under vanilla CE training, and their confidence eventually approaches that of certain samples. In contrast, our method leverages the uncertainty awareness captured by early predictions to prevent overconfident predictions on uncertain samples.

\textbf{Prototype Structure Regularization.} Beyond early predictions, the feature structure of the early model also preserves uncertainty awareness. We further introduce \emph{Prototype Structure Regularization} (PSR) to exploit this uncertainty awareness of the early feature structure. Specifically, given the feature extractor $h_{\boldsymbol{\varphi}^t}$ of the frozen early model, we first construct the prototype for each class. For a sample $\vx_i$, we extract its feature $\vh_i^t$ from the frozen early model, i.e., $\vh_i^t=h_{\boldsymbol{\varphi}^t}(\vx_i)\in\mathbb{R}^d$. Given the uncertainty awareness in early predictions, we utilize the probability assigned by the early model to the true class of each sample as a reliability weight, i.e., $w_i=p_{i,y_i}^t$. Accordingly, the prototype $\vmu_k$ is constructed using normalized confidence-weighted averaging over the early features of class $k$, i.e.,
\begin{equation}
    \vmu_k=\frac{\sum_{i=1}^N\mathbb{I}(y_i=k)w_i\overline{\vh}_i^t}{\sum_{i=1}^N\mathbb{I}(y_i=k)w_i}, 
    \label{eq2}
\end{equation}
where $\mathbb{I}(\cdot)$ is the indicator function and $\overline{\vh}_i^t=\vh_i^t/||\vh_i^t||_2$ is the normalized feature. Given the set of prototypes $\mathcal{M}=\{\vmu_k\}_{k=1}^K$, the similarity between the feature of the sample $\vx_i$ in the early model and the prototype of class $k$ is
\begin{equation}
    s_{i,k}=\cos(\vh_i^t,\vmu_k)=\frac{(\vh_i^t)^\top\vmu_k}{||\vh_i^t||_2||\vmu_k||_2}.
    \label{eq3}
\end{equation}

As observed in Figure \ref{fig1}(b), features near cluster boundaries tend to have lower confidence, whereas those closer to cluster centers tend to have higher confidence. This suggests that similarities to cluster centers, i.e., class prototypes, can reflect the early model's uncertainty awareness in its features. Accordingly, we convert the similarity into a probability distribution over classes for the sample $\vx_i$ using the softmax function, i.e.,
\begin{equation}
    \vq_i=\{q_{i,k}\}_{k=1}^K=\mathrm{softmax}\left(\frac{\vs_i}{\tau}\right), \quad q_{i,k}=\frac{\exp( s_{i,k}/\tau)}{\sum_{j=1}^K\exp( s_{i,j}/\tau)},
    \label{eq4}
\end{equation}
where $\vq_i$ is defined as the prototype prediction of the sample $\vx_i$, $\vs_i=[s_{i,1},s_{i,2},\ldots,s_{i,K}]$ is the similarity vector between the feature of the sample $\vx_i$ and the prototypes, and $\tau>0$ is the scaling factor. The PSR loss is defined as the KL divergence between the prototype prediction distributions and the current prediction distributions, i.e.,
\begin{equation}
    \mathcal{L}_{\mathrm{PSR}}=\frac{1}{N}\sum_{i=1}^ND_{\mathrm{KL}}(\vq_i||\vp_i)=\frac{1}{N}\sum_{i=1}^N\sum_{k=1}^Kq_{i,k}\log\left(\frac{q_{i,k}}{p_{i,k}}\right).
    \label{eq5}
\end{equation}
Note that the prototypes $\{\vmu_k\}_{k=1}^K$ are constructed from the frozen early model, preventing them from drifting during training. Figure \ref{fig3}(b) shows the classification errors and ECE of the prototype predictions for ResNet-50 on CIFAR-10. The ECE of prototype predictions from the vanilla model reaches its minimum at epoch 60 and increases in later stages. In contrast, the ECE of our prototype predictions continues to decrease, demonstrating better uncertainty awareness of the feature structure. The PSR loss leverages this awareness of uncertainty to further improve the calibration performance.

The overall objective of our method is:
\begin{equation}
    \min_{\vtheta}\mathcal{L}_{\mathrm{EUA}\text{-}\mathrm{Cal}}=\min_{\vtheta}\mathcal{L}_{\mathrm{CE}}+\alpha\mathcal{L}_{\mathrm{EPR}}+\beta\mathcal{L}_{\mathrm{PSR}},
    \label{eq6}
\end{equation}
where $\alpha$ and $\beta$ are hyperparameters that control the strengths of EPR loss and PSR loss, respectively. The pseudo-code of our EUA-Cal method is summarized in Algorithm \ref{alg}.

\algrenewcommand\algorithmicrequire{\textbf{Input:}}
\algrenewcommand\algorithmicensure{\textbf{Output:}}

\begin{algorithm}[t]
\caption{EUA-Cal}
\label{alg}
\begin{algorithmic}[1]
\Require Training set $\mathcal{D}=\{(\vx_i,y_i)\}_{i=1}^{N}$, initial model $f_{\vtheta}=h_{\boldsymbol{\varphi}}\mW$, selected epoch $t$, parameters $\alpha$ and $\beta$, scaling factor $\tau$, and total number of epochs $T$.
\Ensure Calibrated model $f_{\vtheta^T}$.
\State Pre-learn $f_{\vtheta}$ with $\mathcal{L}_{\mathrm{CE}}$ until epoch $t$ to obtain the early model $f_{\vtheta^t}$
\State Compute $\vp_i^t=\mathrm{softmax}(f_{\vtheta^t}(\vx_i))$, $\overline{\vh}_i^t=\vh_i^t/\|\vh_i^t\|_2$, and $w_i=p_{i,y_i}^t$ for all $(\vx_i,y_i)\in\mathcal{D}$.
\State Construct the class prototypes $\{\vmu_k\}_{k=1}^{K}$ according to Eq.~(\ref{eq2}).
\State Compute the prototype similarities and distributions $\vq_i$ according to Eq.~(\ref{eq3}) and Eq.~(\ref{eq4}).
\For{epoch from $1$ to $T$}
    \For{each training batch sampled from $\mathcal{D}$}
        \State Compute $\mathcal{L}_{\mathrm{EPR}}$ and $\mathcal{L}_{\mathrm{PSR}}$ according to Eq.~(\ref{eq1}) and Eq.~(\ref{eq5}).
        \State Update $\vtheta$ by minimizing $\mathcal{L}_{\mathrm{EUA}\text{-}\mathrm{Cal}}$ in Eq.~(\ref{eq6}).
    \EndFor
\EndFor
\end{algorithmic}
\end{algorithm}

\section{Experiments}
\subsection{Experimental Setup}
\textbf{Datasets and Models.} We evaluate our method on both image classification and multiple-choice question answering. For image classification, we conduct experiments on CIFAR-10, CIFAR-100 \citep{cifar}, and Tiny-ImageNet \citep{imagenet} datasets with different network architectures, including ResNet-50, ResNet-101 \citep{resnet}, Wide-ResNet-26-10 \citep{wrn}, DenseNet-121 \citep{dense}, and ViT-B/32 \citep{vit}. For multiple-choice question answering, we consider SciQ \citep{sciq}, ARC-Challenge (ARC-C) and ARC-Easy (ARC-E) \citep{arc}, and CommonsenseQA (CSQA) \citep{csqa} datasets. We fine-tune Llama3-8B, Llama3.1-8B \citep{llama3}, and Qwen3-8B-Base \citep{qwen3} using LoRA \citep{lora}.

\textbf{Baselines and Metrics.} For image classification, we compare EUA-Cal with various calibration methods: (\romannumeral 1) the vanilla model trained with CE; (\romannumeral 2) implicit or explicit regularization, including LS \citep{ls}, FL \citep{focal}, CRL \citep{crl}, Distillation \citep{distillation}, MbLS \citep{mbls}, BalCAL \citep{balcal}, and CSAM \citep{csam}; (\romannumeral 3) data augmentation, including Mixup \citep{mixup} and MIT \citep{mit-a}; (\romannumeral 4) architecture-specific methods, including PLP \citep{plp} and MaC-Cal \citep{maccal}. For multiple-choice question answering with fine-tuned LLMs, we evaluate all methods above except BalCAL, PLP, and MaC-Cal since they rely on specific components. We further include (\romannumeral 5) uncertainty-aware methods, including EDL \citep{edl} and IB-EDL \citep{ibedl}. Calibration is evaluated using ACC, ECE \citep{ts}, and adaptive ECE (AECE) \citep{aece}. For out-of-distribution (OOD) detection, we adopt AUROC and FPR95 as metrics.

More implementation details including dataset splitting, training policies, hyperparameters, and evaluation protocols are provided in Appendix \ref{details}.

\definecolor{oursbg}{RGB}{235,245,255}

\begin{table*}[t]
    \centering
    \caption{Performance comparison between different methods with ResNet-50 on CIFAR-10, CIFAR-100, and Tiny-ImageNet datasets. Results are reported as mean $\pm$ standard deviation (\%). The best and second best results are shown in \textbf{bold} and \underline{underlined}, respectively.}
    \label{tb1}
    \setlength{\tabcolsep}{4.0pt}
    \renewcommand{\arraystretch}{1.08}
    \resizebox{\textwidth}{!}{%
    \begin{NiceTabular}{l|ccc|ccc|ccc|c}
        \CodeBefore
        \rowcolor{oursbg}{15}
        \Body
        \toprule
        \multirow{2}{*}{\textbf{Method}}
        & \multicolumn{3}{c}{\textbf{CIFAR-10}}
        & \multicolumn{3}{c}{\textbf{CIFAR-100}}
        & \multicolumn{3}{c}{\textbf{Tiny-ImageNet}}
        & \multirow{2}{*}{\shortstack{\textbf{Avg.}\\\textbf{Rank}}} \\
        & ECE $\downarrow$ & AECE $\downarrow$ & ACC $\uparrow$
        & ECE $\downarrow$ & AECE $\downarrow$ & ACC $\uparrow$
        & ECE $\downarrow$ & AECE $\downarrow$ & ACC $\uparrow$ & \\
        \midrule
        Vanilla
        & 3.36$\pm$0.17 & 3.31$\pm$0.19 & 94.57$\pm$0.15
        & 9.19$\pm$0.16 & 9.07$\pm$0.26 & 76.75$\pm$0.62
        & 5.54$\pm$0.44 & 5.52$\pm$0.43 & 64.45$\pm$0.45 & 10.11 \\
        LS
        & 3.74$\pm$0.24 & 3.93$\pm$0.33 & 94.78$\pm$0.13
        & 3.96$\pm$0.22 & 3.99$\pm$0.24 & 76.77$\pm$0.13
        & 3.78$\pm$0.27 & 3.70$\pm$0.36 & 63.58$\pm$0.67 & 8.78 \\
        FL
        & 1.81$\pm$0.31 & 1.58$\pm$0.27 & 94.40$\pm$0.28
        & 1.77$\pm$0.31 & 1.71$\pm$0.33 & 75.88$\pm$0.35
        & 1.57$\pm$0.11 & 1.48$\pm$0.18 & 62.94$\pm$0.33 & 6.67 \\
        CRL
        & 0.86$\pm$0.09 & \underline{0.61$\pm$0.03} & 93.94$\pm$0.22
        & 5.81$\pm$0.30 & 5.74$\pm$0.27 & 76.87$\pm$0.39
        & 2.99$\pm$0.31 & 3.04$\pm$0.33 & 63.43$\pm$0.33 & 7.67 \\
        Distillation
        & 3.22$\pm$0.52 & 3.27$\pm$0.49 & 94.10$\pm$0.67
        & 7.51$\pm$0.20 & 7.40$\pm$0.16 & 76.64$\pm$0.12
        & 6.04$\pm$0.24 & 6.01$\pm$0.19 & 65.21$\pm$0.20 & 10.22 \\
        MbLS
        & 3.51$\pm$0.23 & 3.48$\pm$0.21 & 94.54$\pm$0.35
        & 8.62$\pm$0.38 & 8.52$\pm$0.37 & 77.41$\pm$0.49
        & \underline{1.42$\pm$0.05} & 1.34$\pm$0.22 & 64.44$\pm$0.48 & 8.00 \\
        BalCAL
        & 2.46$\pm$0.12 & 2.39$\pm$0.11 & 94.38$\pm$0.13
        & 4.03$\pm$0.27 & 3.93$\pm$0.28 & 77.33$\pm$0.31
        & 1.48$\pm$0.14 & 1.52$\pm$0.13 & 65.26$\pm$0.12 & 6.56 \\
        CSAM
        & 1.04$\pm$0.15 & 0.90$\pm$0.19 & \underline{95.36$\pm$0.20}
        & \underline{1.40$\pm$0.39} & \underline{1.30$\pm$0.50} & \textbf{78.82$\pm$0.88}
        & 6.10$\pm$0.13 & 6.12$\pm$0.13 & \textbf{66.54$\pm$0.40} & \underline{4.67} \\
        Mixup
        & 1.74$\pm$0.38 & 2.25$\pm$0.29 & 95.01$\pm$0.37
        & 5.40$\pm$0.62 & 5.34$\pm$0.55 & 77.78$\pm$0.58
        & 3.30$\pm$0.23 & 3.20$\pm$0.15 & 63.58$\pm$0.74 & 6.78 \\
        MIT-A
        & 1.64$\pm$0.06 & 1.54$\pm$0.05 & \textbf{95.57$\pm$0.03}
        & 1.95$\pm$0.11 & 1.87$\pm$0.30 & \underline{78.46$\pm$0.48}
        & 3.66$\pm$0.82 & 3.77$\pm$0.76 & 64.34$\pm$0.89 & 5.22 \\
        PLP
        & 2.35$\pm$0.12 & 2.31$\pm$0.13 & 94.24$\pm$0.19
        & 4.66$\pm$0.38 & 4.64$\pm$0.36 & 74.60$\pm$0.68
        & 2.07$\pm$0.51 & 2.02$\pm$0.51 & 60.85$\pm$1.68 & 9.00 \\
        MaC-Cal
        & \underline{0.83$\pm$0.15} & 0.82$\pm$0.11 & 94.81$\pm$0.05
        & 3.32$\pm$0.14 & 3.14$\pm$0.21 & 76.07$\pm$0.10
        & 1.44$\pm$0.12 & \underline{1.33$\pm$0.23} & 61.34$\pm$0.20 & 5.22 \\
        \textbf{EUA-Cal}
        & \textbf{0.54$\pm$0.12} & \textbf{0.50$\pm$0.09} & 94.63$\pm$0.05
        & \textbf{1.28$\pm$0.11} & \textbf{1.18$\pm$0.21} & 77.66$\pm$0.32
        & \textbf{1.38$\pm$0.17} & \textbf{1.29$\pm$0.11} & \underline{65.31$\pm$0.47} & \textbf{2.00} \\
        \bottomrule
    \end{NiceTabular}%
    }
\end{table*}

\begin{table*}[t]
    \centering
    \caption{Performance comparison between different methods with Llama3-8B on ARC-C, ARC-E, and SciQ datasets. Results are reported as mean $\pm$ standard deviation (\%). The best and second best results are shown in \textbf{bold} and \underline{underlined}, respectively.}
    \label{tb2}
    \setlength{\tabcolsep}{4.0pt}
    \renewcommand{\arraystretch}{1.08}
    \resizebox{\textwidth}{!}{%
    \begin{NiceTabular}{l|ccc|ccc|ccc|c}
        \CodeBefore
        \rowcolor{oursbg}{14}
        \Body
        \toprule
        \multirow{2}{*}{\textbf{Method}}
        & \multicolumn{3}{c}{\textbf{ARC-C}}
        & \multicolumn{3}{c}{\textbf{ARC-E}}
        & \multicolumn{3}{c}{\textbf{SciQ}}
        & \multirow{2}{*}{\shortstack{\textbf{Avg.}\\\textbf{Rank}}} \\
        & ECE $\downarrow$ & AECE $\downarrow$ & ACC $\uparrow$
        & ECE $\downarrow$ & AECE $\downarrow$ & ACC $\uparrow$
        & ECE $\downarrow$ & AECE $\downarrow$ & ACC $\uparrow$ & \\
        \midrule
        Vanilla
        & 17.24$\pm$0.25 & 17.12$\pm$0.31 & 81.31$\pm$0.64
        & 6.54$\pm$0.03 & 6.48$\pm$0.03 & 92.81$\pm$0.12
        & 6.42$\pm$0.38 & 6.10$\pm$0.41 & 93.40$\pm$0.52 & 8.56 \\
        LS
        & 15.23$\pm$0.96 & 15.54$\pm$1.26 & 79.81$\pm$0.96
        & 3.83$\pm$0.17 & 5.44$\pm$0.78 & 91.91$\pm$0.16
        & 3.16$\pm$0.35 & 5.61$\pm$0.44 & 92.83$\pm$0.21 & 8.11 \\
        FL
        & 12.97$\pm$1.31 & 12.85$\pm$0.93 & 80.94$\pm$0.81
        & 4.34$\pm$0.24 & 4.91$\pm$0.24 & 91.93$\pm$0.10
        & 4.49$\pm$0.12 & 4.78$\pm$0.22 & 93.63$\pm$0.06 & 5.89 \\
        CRL
        & 18.17$\pm$0.62 & 18.04$\pm$0.55 & 80.46$\pm$1.19
        & 6.98$\pm$0.84 & 6.88$\pm$0.86 & 91.83$\pm$0.87
        & 6.21$\pm$0.22 & 5.99$\pm$0.27 & 93.27$\pm$0.38 & 10.67 \\
        Distillation
        & 11.99$\pm$0.76 & 12.26$\pm$0.69 & \underline{81.77$\pm$0.50}
        & 3.79$\pm$0.27 & \underline{4.69$\pm$0.45} & \underline{93.00$\pm$0.45}
        & 3.90$\pm$0.13 & 5.03$\pm$0.22 & 93.57$\pm$0.12 & \underline{4.00} \\
        MbLS
        & 12.17$\pm$10.50 & 13.39$\pm$8.18 & 80.97$\pm$0.82
        & 7.62$\pm$0.40 & 7.60$\pm$0.40 & 91.98$\pm$0.32
        & 6.73$\pm$0.09 & 6.71$\pm$0.10 & 93.07$\pm$0.12 & 9.89 \\
        CSAM
        & 13.17$\pm$0.34 & 13.08$\pm$0.45 & \textbf{82.19$\pm$0.69}
        & 5.13$\pm$0.63 & 5.22$\pm$0.43 & \textbf{93.23$\pm$0.65}
        & 5.26$\pm$0.07 & 5.03$\pm$0.05 & 93.63$\pm$0.06 & 5.00 \\
        Mixup
        & 10.53$\pm$0.68 & 10.86$\pm$0.29 & 79.61$\pm$0.48
        & 6.16$\pm$0.24 & 6.32$\pm$0.25 & 91.88$\pm$0.28
        & 5.84$\pm$0.34 & 5.76$\pm$0.30 & 93.67$\pm$0.25 & 7.56 \\
        MIT-A
        & 6.52$\pm$0.39 & \underline{7.10$\pm$0.66} & 81.31$\pm$0.53
        & 5.87$\pm$0.07 & 5.80$\pm$0.08 & 92.75$\pm$0.21
        & 4.47$\pm$0.14 & \underline{4.33$\pm$0.27} & \textbf{94.73$\pm$0.35} & \underline{4.00} \\
        EDL
        & 11.25$\pm$0.59 & 13.25$\pm$0.66 & 80.23$\pm$0.34
        & 2.60$\pm$0.17 & 5.75$\pm$0.16 & 92.51$\pm$0.15
        & 2.86$\pm$0.05 & 6.22$\pm$0.19 & 93.47$\pm$0.15 & 6.67 \\
        IB-EDL
        & \underline{3.98$\pm$0.50} & 9.98$\pm$1.20 & 81.03$\pm$0.47
        & \underline{2.53$\pm$0.54} & 5.82$\pm$0.57 & 92.65$\pm$0.19
        & \underline{2.31$\pm$0.26} & 5.91$\pm$0.59 & 93.30$\pm$0.30 & 5.00 \\
        \textbf{EUA-Cal}
        & \textbf{3.60$\pm$0.34} & \textbf{4.17$\pm$0.60} & 80.94$\pm$0.47
        & \textbf{1.67$\pm$0.19} & \textbf{1.79$\pm$0.29} & 92.72$\pm$0.19
        & \textbf{2.19$\pm$0.24} & \textbf{2.57$\pm$0.49} & \underline{93.90$\pm$0.28} & \textbf{2.22} \\
        \bottomrule
    \end{NiceTabular}%
    }
\end{table*}

\begin{table*}[t]
    \centering
    \caption{Calibration performance of various methods under distribution shift. Llama3-8B is fine-tuned on ARC-E and evaluated on ARC-C, SciQ, and CSQA. Results are reported as mean $\pm$ standard deviation (\%). The best and second best results are shown in \textbf{bold} and \underline{underlined}, respectively.}
    \label{tb3}
    \setlength{\tabcolsep}{4.0pt}
    \renewcommand{\arraystretch}{1.08}
    \resizebox{\textwidth}{!}{%
    \begin{NiceTabular}{l|ccc|ccc|ccc|c}
        \CodeBefore
        \rowcolor{oursbg}{14}
        \Body
        \toprule
        \multirow{2}{*}{\textbf{Method}}
        & \multicolumn{3}{c}{\textbf{ARC-C}}
        & \multicolumn{3}{c}{\textbf{SciQ}}
        & \multicolumn{3}{c}{\textbf{CSQA}}
        & \multirow{2}{*}{\shortstack{\textbf{Avg.}\\\textbf{Rank}}} \\
        & ECE $\downarrow$ & AECE $\downarrow$ & ACC $\uparrow$
        & ECE $\downarrow$ & AECE $\downarrow$ & ACC $\uparrow$
        & ECE $\downarrow$ & AECE $\downarrow$ & ACC $\uparrow$ & \\
        \midrule
        Vanilla
        & 18.25$\pm$0.24 & 18.03$\pm$0.12 & \textbf{79.81$\pm$0.05}
        & 6.72$\pm$0.39 & 6.64$\pm$0.42 & \underline{92.43$\pm$0.35}
        & 25.42$\pm$0.94 & 25.39$\pm$0.99 & 72.07$\pm$0.98 & 7.56 \\
        LS
        & 18.69$\pm$0.26 & 18.80$\pm$0.43 & 76.17$\pm$0.32
        & 4.98$\pm$0.82 & 7.65$\pm$0.20 & 90.90$\pm$0.90
        & 21.82$\pm$1.55 & 21.66$\pm$1.60 & 72.24$\pm$1.44 & 8.11 \\
        FL
        & 16.26$\pm$1.39 & 16.29$\pm$1.33 & 76.99$\pm$1.45
        & 5.47$\pm$1.19 & 6.16$\pm$1.10 & 90.97$\pm$1.32
        & 19.49$\pm$1.33 & 19.38$\pm$1.33 & 72.89$\pm$0.57 & 5.67 \\
        CRL
        & 21.47$\pm$1.05 & 21.41$\pm$0.99 & 75.81$\pm$0.91
        & 8.07$\pm$1.00 & 7.77$\pm$1.18 & 91.20$\pm$1.13
        & 25.01$\pm$0.26 & 25.01$\pm$0.26 & 72.48$\pm$0.12 & 10.00 \\
        Distillation
        & 16.55$\pm$0.78 & 16.53$\pm$0.65 & 77.87$\pm$1.04
        & 5.17$\pm$0.62 & 6.03$\pm$0.78 & 92.00$\pm$0.95
        & 21.84$\pm$0.49 & 21.78$\pm$0.56 & 72.67$\pm$0.58 & 5.78 \\
        MbLS
        & 23.50$\pm$1.08 & 23.33$\pm$1.11 & 75.65$\pm$1.41
        & 8.32$\pm$0.79 & 8.18$\pm$0.84 & 91.37$\pm$1.03
        & 27.86$\pm$1.63 & 27.82$\pm$1.67 & 70.98$\pm$1.77 & 11.67 \\
        CSAM
        & 20.80$\pm$0.43 & 20.63$\pm$0.53 & 77.93$\pm$0.75
        & 7.21$\pm$0.32 & 6.98$\pm$0.29 & 92.37$\pm$0.40
        & 25.33$\pm$1.06 & 25.20$\pm$1.09 & \textbf{73.46$\pm$0.90} & 7.78 \\
        Mixup
        & 17.85$\pm$0.76 & 17.72$\pm$0.69 & 78.50$\pm$0.70
        & 6.14$\pm$0.26 & 6.20$\pm$0.39 & 91.87$\pm$0.47
        & 23.78$\pm$0.65 & 23.72$\pm$0.64 & 71.14$\pm$0.34 & 7.33 \\
        MIT-A
        & 16.81$\pm$0.31 & 16.52$\pm$0.68 & 79.72$\pm$0.50
        & 6.64$\pm$0.28 & 6.50$\pm$0.25 & 91.57$\pm$0.06
        & 21.83$\pm$0.54 & 21.60$\pm$0.69 & \underline{73.41$\pm$0.29} & 5.56 \\
        EDL
        & 13.27$\pm$0.87 & 13.69$\pm$0.77 & 79.66$\pm$0.21
        & \textbf{2.32$\pm$0.83} & \underline{5.48$\pm$0.10} & \textbf{92.73$\pm$0.42}
        & 18.40$\pm$0.63 & 18.07$\pm$0.72 & 72.54$\pm$0.71 & \underline{2.89} \\
        IB-EDL
        & \underline{11.14$\pm$0.12} & \underline{12.02$\pm$0.35} & \underline{79.80$\pm$0.49}
        & 3.17$\pm$0.94 & 5.79$\pm$0.59 & \underline{92.43$\pm$0.15}
        & \underline{16.05$\pm$1.22} & \underline{16.55$\pm$1.02} & 72.18$\pm$1.02 & 3.00 \\
        \textbf{EUA-Cal}
        & \textbf{9.25$\pm$0.50} & \textbf{9.11$\pm$0.42} & 77.59$\pm$0.47
        & \underline{2.50$\pm$0.36} & \textbf{3.12$\pm$0.34} & 92.17$\pm$0.45
        & \textbf{12.48$\pm$0.96} & \textbf{12.26$\pm$0.88} & 73.11$\pm$0.45 & \textbf{2.56} \\
        \bottomrule
    \end{NiceTabular}%
    }
\end{table*}

\begin{figure}[t]
    \centering
    \includegraphics[width=0.98\linewidth]{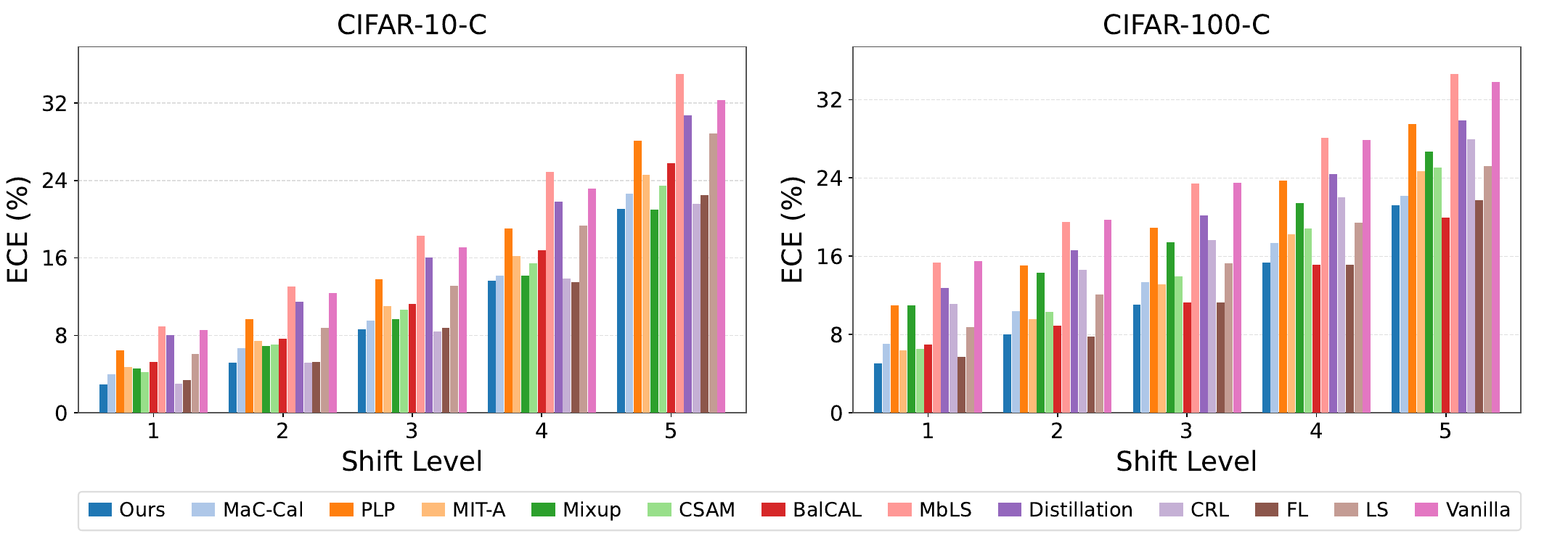}
    \caption{Calibration performance of various methods under distribution shift on CIFAR-10-C and CIFAR-100-C. The horizontal axis denotes the shift level from weakest to strongest and the vertical axis reports ECE (lower is better). }
    \label{fig4}
\end{figure}

\subsection{Calibration Results}

We provide comprehensive comparisons of our method against state-of-the-art methods for model calibration on both image classification and multiple-choice question answering. Table \ref{tb1} reports the calibration performance of our method with ResNet-50 on CIFAR-10, CIFAR-100, and Tiny-ImageNet. Compared with other calibration methods, our method significantly improves calibration performance across all datasets and achieves the best average rank among all methods. Several methods improve calibration at the cost of accuracy, such as the FL and PLP methods. In contrast, our EUA-Cal method obtains the lowest ECE and AECE on all three datasets while maintaining competitive accuracy. These results support the use of uncertainty information retained by the early model for calibration.

Table \ref{tb2} reports the calibration results of Llama3-8B on SciQ, ARC-C, and ARC-E.  Our method also achieves the best average rank and the lowest ECE and AECE on all three datasets. Compared with the second best method on each dataset, our method reduces ECE and AECE by 16.24\% and 47.92\% on average, respectively.  Additional results on ResNet-101, Wide-ResNet-26-10, DenseNet-121, ViT-B/32, Llama3.1-8B, and Qwen3-8B-Base are provided in Appendix \ref{C1}. The consistent improvements across these models further demonstrate the effectiveness and generalizability of EUA-Cal across diverse architectures.

\subsection{Calibration under Distribution Shift}

Maintaining reliable confidence estimates under distribution shift is particularly important in safety-critical applications such as autonomous driving, where systems must make safe decisions even in adverse weather. To evaluate the calibration robustness of our EUA-Cal method in this setting, we first train ResNet-50 on CIFAR-10 and CIFAR-100 and then evaluate on their corrupted versions, CIFAR-10-C and CIFAR-100-C \citep{cifar-c}, respectively. These corruptions include noise, blur, compression artifacts, and changes in brightness. Figure \ref{fig4} illustrates the calibration performance of various methods with different shift levels. Across both datasets and all shift levels, our EUA-Cal method achieves lower ECE than most methods. This indicates that our EUA-Cal method maintains robust calibration performance under distribution shifts. In addition, we also present the OOD detection results in Table \ref{tb4}. Across all settings, our EUA-Cal method consistently outperforms the vanilla model, improving AUROC by an average of 1.31\% and reducing FPR95 by 1.79\%. These results demonstrate that EUA-Cal maintains reliable uncertainty awareness under distribution shift.

Practical deployment of LLMs requires them to generalize beyond the domains covered by their fine-tuning data. To evaluate calibration under such distribution shifts, we fine-tune Llama3-8B on ARC-E and then evaluate on ARC-C, SciQ, and CSQA, as reported in Table~\ref{tb3}. Our method achieves the best average rank and the lowest AECE on all three datasets. It also achieves the lowest ECE on ARC-C and CSQA and ranks second on SciQ. Compared with the second best method on each dataset, our method reduces AECE by 30.94\% on average. These results demonstrate that EUA-Cal maintains effective calibration performance when fine-tuning LLMs under distribution shift.

\begin{table*}[t]
	\centering
	\setlength{\belowcaptionskip}{7pt}
	\caption{OOD detection results (\%). The best and second best results are shown in \textbf{bold} and \underline{underlined}, respectively.}
	\label{tb4}
	\small
	\setlength{\tabcolsep}{3.2pt}
	\renewcommand{\arraystretch}{1}
	\resizebox{0.85\textwidth}{!}{%
		\begin{NiceTabular}{>{\raggedright\arraybackslash}p{\dimexpr0.18\linewidth-2\tabcolsep-0.7pt\relax}|*{4}{>{\centering\arraybackslash}p{\dimexpr0.10\linewidth-2\tabcolsep\relax}}|*{4}{>{\centering\arraybackslash}p{\dimexpr0.10\linewidth-2\tabcolsep\relax}}}
			\CodeBefore
			\rowcolor{oursbg}{9}
			\Body
			\toprule
			\multicolumn{1}{l}{\textsc{ID dataset}} & \multicolumn{4}{c}{\textbf{CIFAR-10}} & \multicolumn{4}{c}{\textbf{Tiny-ImageNet}} \\
			\multicolumn{1}{l}{\textsc{OOD dataset}} & \multicolumn{2}{c}{\textbf{Tiny-ImageNet}} & \multicolumn{2}{c}{\textbf{CIFAR-100}} & \multicolumn{2}{c}{\textbf{CIFAR-10}} & \multicolumn{2}{c}{\textbf{CIFAR-100}} \\
			\cmidrule(lr){2-3}\cmidrule(lr){4-5}\cmidrule(lr){6-7}\cmidrule(lr){8-9}
			\textbf{Method} & AUROC$\uparrow$ & FPR95$\downarrow$ & AUROC$\uparrow$ & FPR95$\downarrow$ & AUROC$\uparrow$ & FPR95$\downarrow$ & AUROC$\uparrow$ & FPR95$\downarrow$ \\
			\midrule
			Vanilla & 88.41 & 58.30 & 87.41 & 60.91 & \underline{85.01} & \underline{67.25} & \underline{83.61} & \underline{68.40} \\
			MbLS & 88.97 & 58.91 & 87.37 & 63.08 & 81.57 & 74.82 & 80.16 & 75.47 \\
			BalCAL & 87.43 & 56.97 & 86.67 & 60.71 & 83.34 & 71.71 & 82.32 & 71.60 \\
			PLP & 88.95 & 60.63 & 88.24 & 63.56 & 80.94 & 81.21 & 80.37 & 79.44 \\
			MaC-Cal & \underline{89.02} & \textbf{55.78} & \underline{88.56} & \textbf{57.30} & 83.27 & 74.60 & 82.29 & 74.24 \\
			\textbf{EUA-Cal} & \textbf{89.32} & \underline{56.09} & \textbf{89.16} & \underline{59.85} & \textbf{86.45} & \textbf{64.83} & \textbf{84.75} & \textbf{66.94} \\
			\bottomrule
		\end{NiceTabular}%
	}
\end{table*}

\begin{table*}[t]
	\centering
	\caption{Ablation study of EPR, PSR, and the anchor model. The best and second best results are shown in \textbf{bold} and \underline{underlined}, respectively.}
	\label{tb:ablation}
	\setlength{\tabcolsep}{4.0pt}
	\renewcommand{\arraystretch}{1}
	\resizebox{\textwidth}{!}{%
		\begin{NiceTabular}{ccc|cccccc|cccccc}
			\CodeBefore
			\rowcolor{oursbg}{8}
			\Body
			\toprule
			\multirow{3}{*}{\textbf{EPR}}
			& \multirow{3}{*}{\textbf{PSR}}
			& \multirow{3}{*}{\shortstack{\textbf{Anchor}\\\textbf{Model}}}
			& \multicolumn{6}{c}{\textbf{ResNet-50}}
			& \multicolumn{6}{c}{\textbf{Llama3-8B}} \\
			& &
			& \multicolumn{2}{c}{\textbf{CIFAR-10}}
			& \multicolumn{2}{c}{\textbf{CIFAR-100}}
			& \multicolumn{2}{c}{\textbf{Tiny-ImageNet}}
			& \multicolumn{2}{c}{\textbf{ARC-C}}
			& \multicolumn{2}{c}{\textbf{ARC-E}}
			& \multicolumn{2}{c}{\textbf{SciQ}} \\
			& &
			& ECE $\downarrow$ & ACC $\uparrow$
			& ECE $\downarrow$ & ACC $\uparrow$
			& ECE $\downarrow$ & ACC $\uparrow$
			& ECE $\downarrow$ & ACC $\uparrow$
			& ECE $\downarrow$ & ACC $\uparrow$
			& ECE $\downarrow$ & ACC $\uparrow$ \\
			\midrule
			$\times$ & $\times$ & --
			& 3.36 & 94.57
			& 9.19 & 76.75
			& 5.54 & 64.45
			& 17.24 & \textbf{81.31}
			& 6.54 & \underline{92.81}
			& 6.42 & 93.40 \\
			$\checkmark$ & $\times$ & Early
			& \underline{0.75} & \underline{94.69}
			& 3.30 & 77.57
			& \underline{2.50} & \underline{65.48}
			& \underline{7.47} & \underline{81.10}
			& \underline{2.67} & 91.25
			& 4.21 & 93.30 \\
			$\times$ & $\checkmark$ & Early
			& 0.83 & 94.67
			& \underline{2.23} & 77.37
			& 6.62 & 64.41
			& 7.79 & 81.07
			& 4.67 & 91.33
			& \underline{2.44} & 93.10 \\
			$\checkmark$ & $\checkmark$ & Late
			& 2.14 & \textbf{94.78}
			& 6.24 & \textbf{78.11}
			& 3.63 & \textbf{66.15}
			& 9.71 & 80.12
			& 6.64 & \textbf{92.85}
			& 5.11 & \textbf{94.30} \\
			$\checkmark$ & $\checkmark$ & Early
			& \textbf{0.54} & 94.63
			& \textbf{1.28} & \underline{77.66}
			& \textbf{1.38} & 65.31
			& \textbf{3.60} & 80.94
			& \textbf{1.67} & 92.72
			& \textbf{2.19} & \underline{93.90} \\
			\bottomrule
		\end{NiceTabular}%
	}
\end{table*}

\begin{figure}[!t]
	\centering
	
	\subfloat[ResNet-50 on CIFAR-10]{%
		\includegraphics[width=0.51\linewidth]{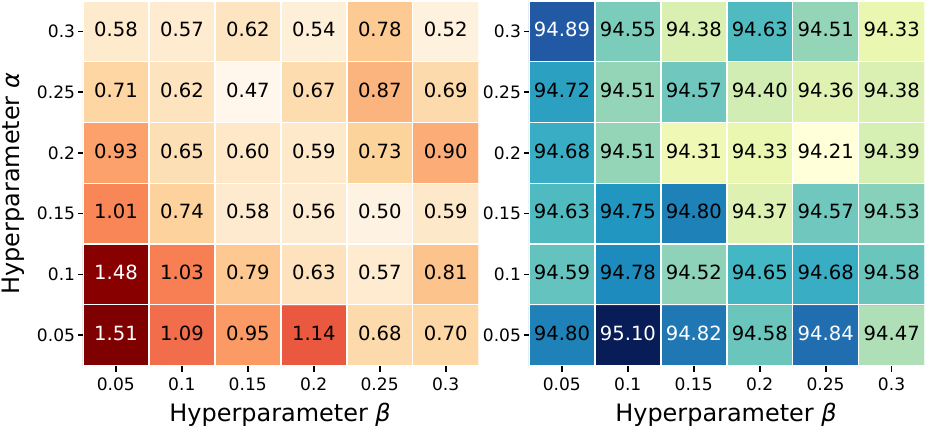}
		\label{5a}
	}
	\subfloat[Llama3-8B on ARC-E]{%
		\includegraphics[width=0.49\linewidth]{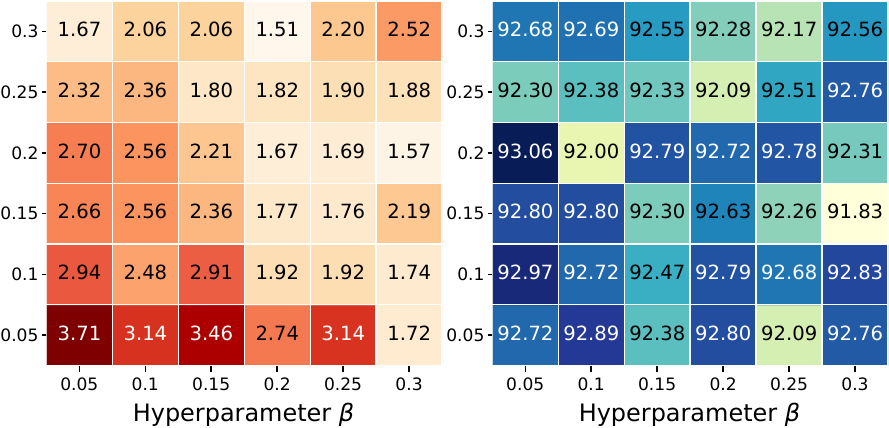}
		\label{5b}
	}
	\caption{Sensitivity to $\alpha$ and $\beta$ of EUA-Cal. ECE ($\downarrow$) is shown in \textcolor{orange}{warmer colors}, and ACC ($\uparrow$) is shown in \textcolor{blue}{cooler colors}.}
	\label{fig5}
\end{figure}

\subsection{Further Analysis}
\textbf{Ablation Study.} As shown in Table \ref{tb:ablation}, we evaluate several variants of our EUA-Cal method on ResNet-50 and Llama3-8B to investigate the effectiveness of EPR, PSR, and the selection of the anchor model. We find that the EPR variant reduces the ECE across all six datasets, while the PSR variant also improves calibration on most datasets. The complete EUA-Cal, which combines the two regularization terms, achieved the lowest ECE on each dataset, suggesting that predictions and feature structures of the early model provide complementary uncertainty awareness for calibration. For anchor model selection, we compare the early model with a late model that has higher validation accuracy but also higher validation ECE. Although the late model leads to slightly higher accuracy in most settings, its calibration is substantially worse. More analysis on the impact of model selection is provided in Appendix~\ref{C2}.

\textbf{Impact of Hyperparameters.} We explore the impact of hyperparameters $\alpha$, $\beta$ and $\tau$ on ResNet-50 and Llama3-8B. $\alpha$ and $\beta$ control the strengths of EPR loss and PSR loss, respectively. We recommend setting them within the range of $[0.1,0.3]$. As shown in Figure \ref{fig5}, our EUA-Cal method consistently improves the calibration performance within this range. Figure~\ref{fig9} shows the performance of EUA-Cal under different values of $\tau$, where $\tau^*$ is obtained by minimizing NLL on the training set. Our EUA-Cal method maintains relatively stable accuracy and competitive calibration performance under different $\tau$ settings. We recommend determining $\tau$ through minimizing NLL on the training set, avoiding a separate hyperparameter search. For more detailed analyses, see Appendix \ref{C3}.

\section{Conclusion}
In this work, we observe a consistent phenomenon across different models that the early model is better calibrated, while continued training often yields modest accuracy gains while worsening calibration. Motivated by this observation, we proposed EUA-Cal, a novel method that exploits the early model as an uncertainty anchor for calibration. Our method is achieved through the regularization terms EPR and PSR to leverage the uncertainty awareness of the early model for calibration. Extensive experiments on image classification and multiple-choice question answering on eight models indicate that our EUA-Cal method consistently improves calibration performance while maintaining competitive accuracy across different models and tasks.

\subsection*{AI use statement}

In this work, we used generative AI tools solely to assist with grammar, spelling, and sentence clarity. We have reviewed all AI-assisted work and assume full responsibility for all content generated by the LLM. The generative AI contributed in no other way to this work. The research idea, experiment design, and all other content were developed and completed by ourselves.

\subsection*{Ethics statement}

We confirm that our work adheres to the ICLR Code of Ethics. The research utilizes publicly available datasets that contain no personally identifiable information. As a strictly technical contribution, we foresee no ethical issues or potential misuse.

\subsection*{Reproducibility statement}

The complete code for our proposed method is provided in the Supplementary Material to facilitate the reproduction of our results. Our experimental implementation is also described in Appendix~\ref{details}.

\bibliography{iclr2027_conference}
\bibliographystyle{iclr2027_conference}

\appendix
\clearpage

\section{Implementation Details}
\label{details}

\subsection{Datasets}
To comprehensively evaluate the effectiveness of our method, we adopt CIFAR-10, CIFAR-100 \citep{cifar}, and Tiny-ImageNet \citep{imagenet} for image classification, and SciQ \citep{sciq}, ARC-C and ARC-E \citep{arc}, and CommonsenseQA (CSQA) \citep{csqa} for multiple-choice question answering. 

\textbf{CIFAR-10 and CIFAR-100} each contain 60,000 RGB images of size $32\times32$, covering 10 and 100 classes, respectively. For each dataset, we
divide the original 50,000 training images into 45,000 training and 5,000 validation images, and retain the 10,000 test images. 

\textbf{Tiny-ImageNet} contains 200 classes of natural images at $64\times64$ resolution. We divide its 100,000 training images into 90,000 for training and 10,000 for validation, and evaluate on the labeled test set of 10,000 images. 

\textbf{SciQ} contains 13,679 four-choice science questions that cover physics, chemistry, biology, and related subjects. We use the official training/validation/test splits of 11,679/1,000/1,000 examples. During fine-tuning, each question and its answer choices are provided as input, while the correct answer serves as the supervision label. 

\textbf{ARC-C and ARC-E} consist of grade-school science examination questions. Specifically, ARC-C contains challenging questions that retrieval-based and word co-occurrence baselines fail to answer, while ARC-E contains the remaining questions. We follow the official training/validation/test splits of 1,119/299/1,172 for ARC-C and 2,251/570/2,376 for ARC-E.

\textbf{CommonsenseQA} contains 12,102 five-choice questions requiring common sense knowledge about everyday situations. Its official training/development/test splits contain 9,741/1,221/1,140 examples. Since test labels are unavailable, we use the labeled development set for evaluation.

\subsection{Training Details}
For image classification, we conduct experiments on five models including ResNet-50, ResNet-101, Wide-ResNet-26-10, DenseNet-121, and ViT-B/32. To learn these models, we use SGD as the optimizer with a momentum of 0.9 and a weight decay of $5\times10^{-4}$ and an initial learning rate of 0.1 unless otherwise specified. On CIFAR-10 and CIFAR-100, models are trained for 200 epochs with a batch size of 128. The learning rate is divided by 5 at epochs 60, 120, and 160. On Tiny-ImageNet, models are trained for 100 epochs with a batch size of 64. The learning rate is divided by 10 at epochs 40 and 60. We initialize ViT-B/32 from ImageNet-1K pretrained weights and fine-tune all its parameters using $224\times224$ inputs. For ViT-B/32 on CIFAR, we use a batch size of 64, an initial learning rate of 0.01 with cosine decay, no weight decay, and a gradient clipping norm of 1.

For Multiple-choice question answering, we fine-tune Llama3-8B, Llama3.1-8B, and Qwen3-8B-Base using LoRA. Following \citep{ibedl}, we define the label space $\mathcal{Y}$ as the tokens corresponding to the possible options (A/B/C/D). The adapters are applied to the query, key, value, and output projections, with rank 8, scaling factor 16, and dropout 0.1. The learning rate is set to $5\times10^{-5}$ and annealed using a cosine schedule. The maximum token length is limited to 256 for all other datasets. The training is conducted with bfloat16 precision.

\subsection{Hyperparameter Settings}
We compare our EUA-Cal method with various calibration methods and provide hyperparameter settings of these baseline methods as follows. (a) LS \citep{ls}: We use a smoothing factor of $\alpha=0.05$. (b) FL \citep{focal}: We set the focusing parameter to $\gamma=2$. (c) CRL \citep{crl}: We set the coefficient of the confidence ranking loss to $1.0$. (d) Distillation \citep{distillation}: We use two teachers and set the distillation loss weight to $1.0$. The distillation temperature is determined by minimizing the validation Brier score. (e) MbLS \citep{mbls}: We use a margin of $m=10$ and a regularization coefficient of $0.1$. (f) BalCAL \citep{balcal}: We set $\beta=1,\delta=0.95$ with a dynamic $\gamma$. (g) CSAM \citep{csam}: We use a perturbation radius $\rho\in[0.05,0.2]$ and $\gamma\in[0.5,2.0]$. (h) Mixup \citep{mixup}: We use $\alpha=0.2$. (i) MIT \citep{mit-a}: We adopt MIT-A with $\alpha=0.5$ and $\Delta\lambda>1/2$. (j) PLP\citep{plp}: We set the hyperparameter $\gamma$ to $1.0$.(k) MaC-Cal \citep{maccal}: We set the weighting factor $\gamma$ to $0.7$. (l) EDL \citep{edl}: We limit the annealing coefficient to $1.0$. (m) IB-EDL \citep{ibedl}: We use $K=20$ samples and $\beta\in[5\times10^{-7},9\times10^{-5}]$, depending on the dataset.

For our EUA-Cal experiments, we choose $\alpha$ and $\beta$ from $[0.1,0.3]$ and determine $\tau$ by minimizing the NLL of prototype predictions on training samples. The anchor is selected at epoch 60 for CIFAR-10 and CIFAR-100 and epoch 40 for Tiny-ImageNet, except for ViT-B/32, which uses epoch 5. In LLM experiments, we use either the pretrained model or a model obtained after 5-15 fine-tuning steps as the anchor, depending on the model and dataset.

\begin{figure}[t]
    \centering

    \subfloat[Llama3-8B on SciQ]{%
        \includegraphics[width=0.25\linewidth]{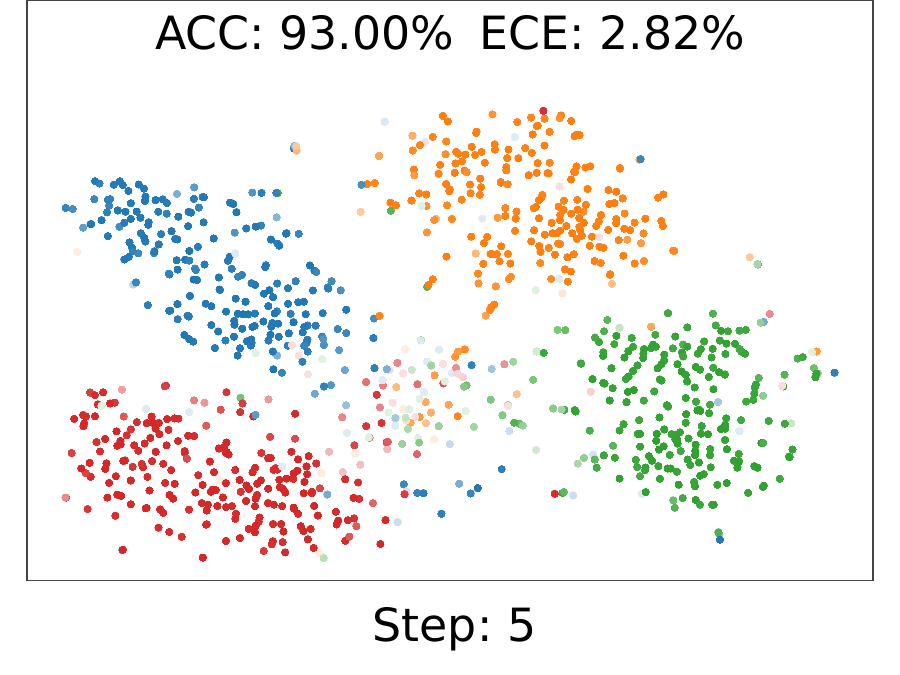}%
        \includegraphics[width=0.25\linewidth]{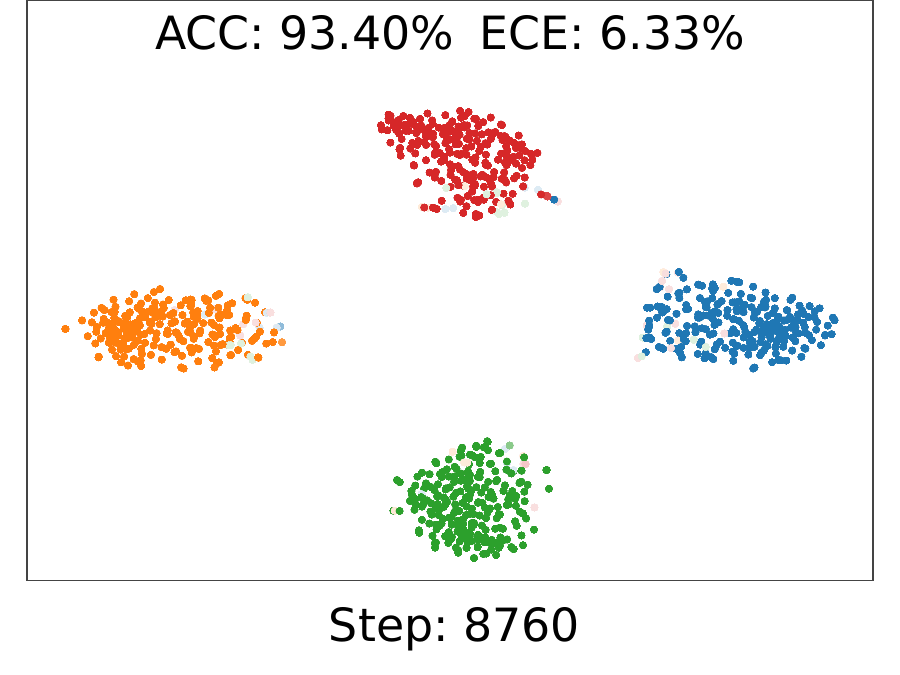}%
        \label{6b}
    }
    \subfloat[Qwen3-8B-Base on SciQ]{%
        \includegraphics[width=0.25\linewidth]{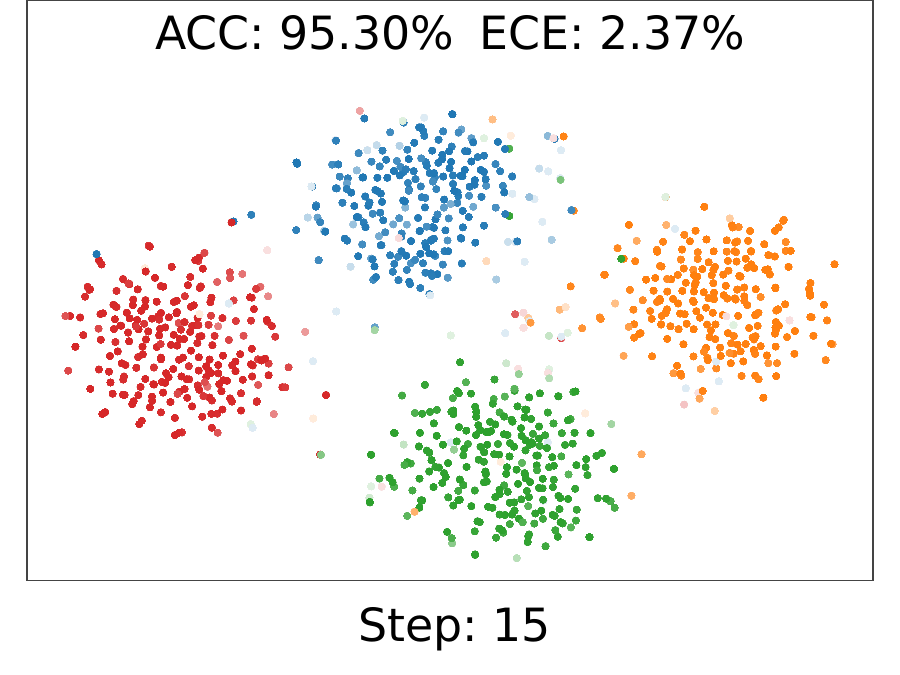}%
        \includegraphics[width=0.25\linewidth]{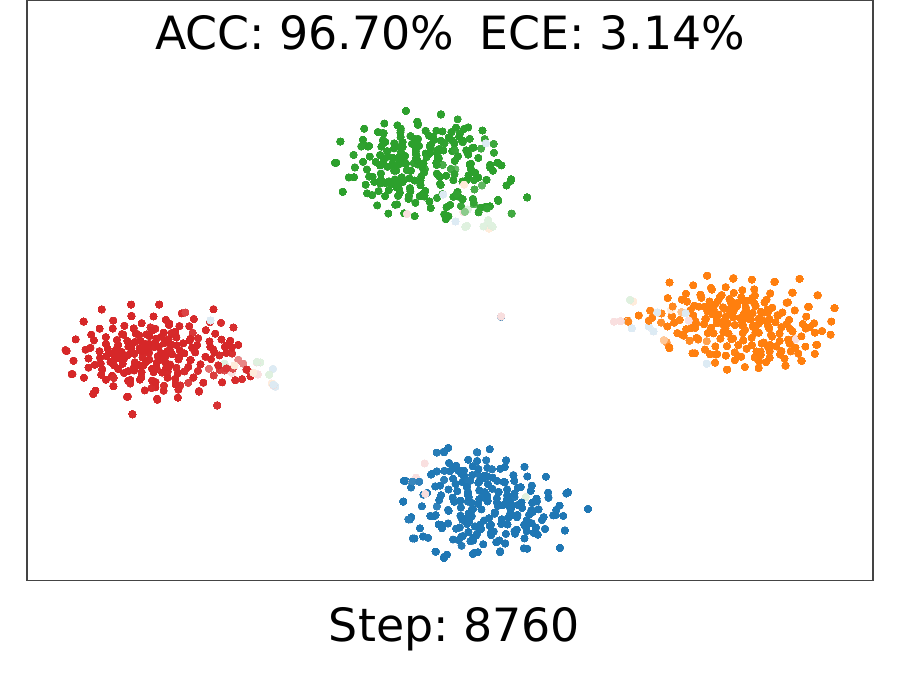}%
        \label{6b}
    }
    \caption{t-SNE visualizations of features from Llama3-8B and Qwen3-8B-Base on SciQ at different fine-tuning steps. Lighter shades indicate features associated with lower confidence.} 
    \label{fig6}
\end{figure}

\begin{table*}[t]
    \centering
    \caption{Performance comparison between different methods with ResNet-101 on CIFAR-10, CIFAR-100, and Tiny-ImageNet datasets. Results are reported as mean $\pm$ standard deviation (\%). The best and second best results are shown in \textbf{bold} and \underline{underlined}, respectively.}
    \label{tb6}
    \setlength{\tabcolsep}{4.0pt}
    \renewcommand{\arraystretch}{1.08}
    \resizebox{\textwidth}{!}{%
    \begin{NiceTabular}{l|ccc|ccc|ccc|c}
        \CodeBefore
        \rowcolor{oursbg}{15}
        \Body
        \toprule
        \multirow{2}{*}{\textbf{Method}}
        & \multicolumn{3}{c}{\textbf{CIFAR-10}}
        & \multicolumn{3}{c}{\textbf{CIFAR-100}}
        & \multicolumn{3}{c}{\textbf{Tiny-ImageNet}}
        & \multirow{2}{*}{\shortstack{\textbf{Avg.}\\\textbf{Rank}}} \\
        & ECE $\downarrow$ & AECE $\downarrow$ & ACC $\uparrow$
        & ECE $\downarrow$ & AECE $\downarrow$ & ACC $\uparrow$
        & ECE $\downarrow$ & AECE $\downarrow$ & ACC $\uparrow$ & \\
        \midrule
        Vanilla
        & 3.26$\pm$0.13 & 3.21$\pm$0.12 & 94.81$\pm$0.16
        & 8.97$\pm$0.28 & 8.94$\pm$0.27 & 78.21$\pm$0.67
        & 5.95$\pm$0.39 & 5.99$\pm$0.36 & 65.74$\pm$0.26 & 9.78 \\
        LS
        & 4.05$\pm$0.27 & 4.21$\pm$0.18 & 94.72$\pm$0.38
        & 3.81$\pm$0.47 & 4.01$\pm$0.40 & 77.57$\pm$0.48
        & 1.90$\pm$0.05 & 1.69$\pm$0.17 & 66.22$\pm$0.38 & 8.33 \\
        FL
        & 1.63$\pm$0.14 & 1.42$\pm$0.09 & 94.68$\pm$0.26
        & \underline{1.57$\pm$0.26} & \underline{1.61$\pm$0.19} & 77.03$\pm$0.61
        & 1.38$\pm$0.08 & \underline{1.29$\pm$0.19} & 63.68$\pm$0.29 & 6.11 \\
        CRL
        & \underline{0.83$\pm$0.13} & \underline{0.67$\pm$0.05} & 94.10$\pm$0.15
        & 5.91$\pm$0.24 & 5.78$\pm$0.06 & 77.36$\pm$0.50
        & 3.57$\pm$0.19 & 3.49$\pm$0.24 & 64.73$\pm$0.28 & 8.33 \\
        Distillation
        & 2.87$\pm$0.15 & 2.95$\pm$0.14 & 94.86$\pm$0.04
        & 7.17$\pm$0.39 & 7.12$\pm$0.35 & 77.93$\pm$0.26
        & 7.36$\pm$0.14 & 7.36$\pm$0.13 & 66.04$\pm$0.15 & 9.33 \\
        MbLS
        & 3.57$\pm$0.08 & 3.58$\pm$0.10 & 94.71$\pm$0.02
        & 9.21$\pm$0.53 & 9.18$\pm$0.48 & 77.84$\pm$0.53
        & 1.49$\pm$0.30 & 1.34$\pm$0.16 & 65.68$\pm$0.56 & 9.44 \\
        BalCAL
        & 2.23$\pm$0.20 & 2.08$\pm$0.18 & 94.75$\pm$0.07
        & 3.56$\pm$0.13 & 3.40$\pm$0.09 & 78.17$\pm$0.35
        & 1.32$\pm$0.22 & 1.44$\pm$0.23 & \underline{66.59$\pm$0.50} & 5.44 \\
        CSAM
        & 1.36$\pm$0.08 & 1.28$\pm$0.07 & \underline{95.53$\pm$0.14}
        & 1.64$\pm$0.26 & 1.70$\pm$0.25 & \textbf{79.60$\pm$0.06}
        & 5.72$\pm$0.26 & 5.70$\pm$0.30 & \textbf{67.12$\pm$0.42} & \underline{4.44} \\
        Mixup
        & 1.61$\pm$0.41 & 2.21$\pm$0.06 & 95.47$\pm$0.20
        & 4.93$\pm$0.83 & 4.88$\pm$0.85 & 78.52$\pm$0.49
        & 1.89$\pm$0.54 & 1.79$\pm$0.70 & 65.19$\pm$0.84 & 6.44 \\
        MIT-A
        & 1.66$\pm$0.19 & 1.60$\pm$0.18 & \textbf{95.87$\pm$0.19}
        & 2.07$\pm$0.57 & 2.05$\pm$0.54 & \underline{79.13$\pm$0.46}
        & 2.84$\pm$0.73 & 2.76$\pm$0.76 & 65.31$\pm$0.42 & 5.56 \\
        PLP
        & 2.25$\pm$0.17 & 2.22$\pm$0.17 & 94.04$\pm$0.37
        & 8.76$\pm$0.34 & 8.75$\pm$0.34 & 75.74$\pm$0.45
        & 2.23$\pm$0.36 & 2.06$\pm$0.47 & 59.49$\pm$2.05 & 10.56 \\
        MaC-Cal
        & 1.18$\pm$0.11 & 1.25$\pm$0.14 & 94.94$\pm$0.03
        & 3.97$\pm$0.18 & 3.80$\pm$0.20 & 78.31$\pm$0.07
        & \underline{1.27$\pm$0.26} & 1.30$\pm$0.16 & 63.22$\pm$0.27 & 5.00 \\
        \textbf{EUA-Cal}
        & \textbf{0.52$\pm$0.10} & \textbf{0.49$\pm$0.09} & 94.74$\pm$0.14
        & \textbf{1.07$\pm$0.27} & \textbf{1.03$\pm$0.28} & 78.58$\pm$0.19
        & \textbf{1.07$\pm$0.10} & \textbf{1.11$\pm$0.14} & 66.24$\pm$0.08 & \textbf{2.22} \\
        \bottomrule
    \end{NiceTabular}%
    }
\end{table*}

\subsection{Evaluation Protocols}

We report expected calibration error (ECE) and adaptive expected calibration error (AECE) to evaluate confidence calibration. ECE measures the difference between accuracy and confidence in expectation, i.e., $\mathbb{E}[|P(\hat{y}_i=y_i|\hat{p}_i)-\hat{p}_i|]$. Dividing the confidence interval $[0,1]$ into $M$ equally spaced bins $\{B_m\}^M_{m=1}$,  ECE is defined as the weighted average of the difference between accuracy and confidence in each bin, i.e., 
\begin{equation}
\begin{aligned}
\mathrm{ECE}
&=\sum_{m=1}^{M}\frac{|B_m|}{N}
\left|\mathrm{acc}(B_m)-\mathrm{conf}(B_m)\right|,
\end{aligned}
\end{equation}

where $N$ is the total number of samples, $|B_m|$ is the number of samples in bin $m$, $\mathrm{acc}(B_m)$ and $\mathrm{conf}(B_m)$ denote the accuracy and the average confidence of the samples in this bin, respectively. Unlike ECE, AECE adaptively divides the confidence interval with the same sample size. Lower ECE and AECE indicate better calibration performance. In our experiments, we set the number of bins as 20 for both ECE and AECE. The results are reported as the mean and standard deviation over three random seeds.

\section{Visualizations}
\label{vis}
Figure~\ref{fig6} illustrates a similar pattern during LLM fine-tuning. For both models, the later features form more distinct class clusters, while the number of the low-confidence samples visible at the early steps decreases. Accuracy increases from 93.00\% to 93.40\% for Llama3-8B and from 95.30\% to 96.70\% for Qwen3-8B-Base, but ECE also rises from 2.82\% to 6.33\% and from 2.37\% to 3.14\%, respectively. This indicates that the uncertainty awareness of the features preserved by the early model is lost in the later training stage.

\section{More Experimental Results}
\label{moreResuts}

\subsection{Calibration on Additional Model Architecture}
\label{C1}

Tables~\ref{tb6}--\ref{tb9} report additional image classification results on ResNet-101, Wide-ResNet-26-10, DenseNet-121, and ViT-B/32. Our EUA-Cal method achieves the lowest ECE on every dataset for all four models and the lowest AECE in eight of the nine cases while maintaining competitive accuracy. In particular, on ViT-B/32, it reduces ECE from 1.92\% to 0.48\% on CIFAR-10 and from 9.56\% to 0.89\% on CIFAR-100 compared with the vanilla model. Table~\ref{tb10} and Table~\ref{tb11} present multiple-choice question answering results on Llama3.1-8B and Qwen3-8B-Base, respectively. Our EUA-Cal method achieves the best average rank on both models, with the lowest AECE in five of six settings. These results further demonstrate its effectiveness across diverse models and tasks.

\begin{table*}[t]
    \centering
    \caption{Performance comparison between different methods with Wide-ResNet-26-10 on CIFAR-10 and CIFAR-100. Results are reported as mean $\pm$ standard deviation (\%). The best and second best results are shown in \textbf{bold} and \underline{underlined}, respectively.}
    \label{tb7}
    \setlength{\tabcolsep}{4.0pt}
    \renewcommand{\arraystretch}{1.08}
    \resizebox{0.8\textwidth}{!}{%
    \begin{NiceTabular}{l|ccc|ccc}
        \CodeBefore
        \rowcolor{oursbg}{9}
        \Body
        \toprule
        \multirow{2}{*}{\textbf{Method}}
        & \multicolumn{3}{c}{\textbf{CIFAR-10}}
        & \multicolumn{3}{c}{\textbf{CIFAR-100}}\\
        & ECE $\downarrow$ & AECE $\downarrow$ & ACC $\uparrow$
        & ECE $\downarrow$ & AECE $\downarrow$ & ACC $\uparrow$  \\
        \midrule
        Vanilla
        & 2.52$\pm$0.07 & 2.43$\pm$0.05 & 95.84$\pm$0.08
        & 5.41$\pm$0.12 & 5.26$\pm$0.17 & 79.64$\pm$0.19  \\
        MbLS
        & 2.79$\pm$0.10 & 2.75$\pm$0.12 & 95.64$\pm$0.11
        & 7.48$\pm$0.11 & 7.32$\pm$0.09 & 79.66$\pm$0.15  \\
        BalCAL
        & 2.57$\pm$0.24 & 2.40$\pm$0.12 & 95.69$\pm$0.07
        & 4.76$\pm$0.20 & 4.53$\pm$0.33 & 78.39$\pm$0.58  \\
        CSAM
        & \underline{0.70$\pm$0.07} & \textbf{0.54$\pm$0.07} & \underline{96.52$\pm$0.08}
        & \underline{1.95$\pm$0.16} & \underline{1.84$\pm$0.16} & \textbf{81.70$\pm$0.13} \\
        MIT-A
        & 1.30$\pm$0.11 & 1.28$\pm$0.10 & \textbf{96.70$\pm$0.12}
        & 2.98$\pm$0.15 & 2.89$\pm$0.20 & \underline{81.30$\pm$0.12} \\
        MaC-Cal
        & 1.18$\pm$0.16 & 1.25$\pm$0.03 & 95.93$\pm$0.04
        & 2.70$\pm$0.10 & 2.55$\pm$0.03 & 78.87$\pm$0.11  \\
        \textbf{EUA-Cal}
        & \textbf{0.59$\pm$0.12} & \underline{0.58$\pm$0.14} & 95.79$\pm$0.07
        & \textbf{1.72$\pm$0.12} & \textbf{1.61$\pm$0.15} & 80.95$\pm$0.04  \\
        \bottomrule
    \end{NiceTabular}%
    }
\end{table*}

\begin{table*}[t]
    \centering
    \caption{Performance comparison between different methods with DenseNet-121 on CIFAR-10 and CIFAR-100. Results are reported as mean $\pm$ standard deviation (\%). The best and second best results are shown in \textbf{bold} and \underline{underlined}, respectively.}
    \label{tb8}
    \setlength{\tabcolsep}{4.0pt}
    \renewcommand{\arraystretch}{1.08}
    \resizebox{0.8\textwidth}{!}{%
    \begin{NiceTabular}{l|ccc|ccc}
        \CodeBefore
        \rowcolor{oursbg}{9}
        \Body
        \toprule
        \multirow{2}{*}{\textbf{Method}}
        & \multicolumn{3}{c}{\textbf{CIFAR-10}}
        & \multicolumn{3}{c}{\textbf{CIFAR-100}}\\
        & ECE $\downarrow$ & AECE $\downarrow$ & ACC $\uparrow$
        & ECE $\downarrow$ & AECE $\downarrow$ & ACC $\uparrow$  \\
        \midrule
        Vanilla
        & 2.92$\pm$0.18 & 2.87$\pm$0.18 & \underline{94.69$\pm$0.19}
        & 8.38$\pm$0.26 & 8.34$\pm$0.25 & 76.11$\pm$0.25  \\
        MbLS
        & 2.96$\pm$0.04 & 2.92$\pm$0.07 & 94.25$\pm$0.20
        & 2.38$\pm$0.33 & 2.12$\pm$0.33 & 75.95$\pm$0.44 \\
        BalCAL
        & 1.07$\pm$0.02 & 0.99$\pm$0.08 & 93.96$\pm$0.41
        & 7.93$\pm$0.49 & 7.93$\pm$0.49 & 74.41$\pm$0.52  \\
        CSAM
        & 4.33$\pm$0.12 & 4.32$\pm$0.10 & 94.50$\pm$0.07
        & 4.42$\pm$0.31 & 4.40$\pm$0.31 & \underline{77.26$\pm$0.21}  \\
        MIT-A
        & 0.90$\pm$0.04 & 0.78$\pm$0.01 & \textbf{95.87$\pm$0.12}
        & 2.77$\pm$0.10 & 2.40$\pm$0.22 & \textbf{78.57$\pm$0.31}  \\
        MaC-Cal
        & \underline{0.83$\pm$0.07} & \underline{0.70$\pm$0.03} & 94.28$\pm$0.07
        & \underline{1.41$\pm$0.07} & \underline{1.25$\pm$0.06} & 70.92$\pm$0.62 \\
        \textbf{EUA-Cal}
        & \textbf{0.55$\pm$0.11} & \textbf{0.40$\pm$0.08} & 94.34$\pm$0.09
        & \textbf{1.13$\pm$0.21} & \textbf{1.03$\pm$0.26} & 76.27$\pm$0.18  \\
        \bottomrule
    \end{NiceTabular}%
    }
\end{table*}

\begin{table*}[t]
    \centering
    \caption{Performance comparison between different methods with ViT-B/32 on CIFAR-10 and CIFAR-100. Results are reported as mean $\pm$ standard deviation (\%). The best and second best results are shown in \textbf{bold} and \underline{underlined}, respectively.}
    \label{tb9}
    \setlength{\tabcolsep}{4.0pt}
    \renewcommand{\arraystretch}{1.08}
    \resizebox{0.8\textwidth}{!}{%
    \begin{NiceTabular}{l|ccc|ccc}
        \CodeBefore
        \rowcolor{oursbg}{9}
        \Body
        \toprule
        \multirow{2}{*}{\textbf{Method}}
        & \multicolumn{3}{c}{\textbf{CIFAR-10}}
        & \multicolumn{3}{c}{\textbf{CIFAR-100}}\\
        & ECE $\downarrow$ & AECE $\downarrow$ & ACC $\uparrow$
        & ECE $\downarrow$ & AECE $\downarrow$ & ACC $\uparrow$ \\
        \midrule
        Vanilla
        & 1.92$\pm$0.10 & 1.90$\pm$0.10 & 97.78$\pm$0.10
        & 9.56$\pm$0.06 & 9.54$\pm$0.07 & 87.61$\pm$0.07 \\
        MbLS
        & 1.80$\pm$0.09 & 1.77$\pm$0.11 & \underline{97.90$\pm$0.10}
        & 8.65$\pm$0.19 & 8.58$\pm$0.19 & \underline{87.71$\pm$0.16} \\
        BalCAL
        & 3.88$\pm$0.08 & 4.05$\pm$0.14 & 97.67$\pm$0.07
        & 4.83$\pm$0.16 & 5.60$\pm$0.21 & 87.49$\pm$0.19 \\
        CSAM
        & 1.52$\pm$0.06 & 1.49$\pm$0.05 & 97.56$\pm$0.02
        & 5.32$\pm$0.03 & 5.28$\pm$0.04 & 87.26$\pm$0.01  \\
        MIT-L
        & 1.20$\pm$0.02 & 1.13$\pm$0.04 & \textbf{98.13$\pm$0.05}
        & 6.67$\pm$0.06 & 6.65$\pm$0.05 & \textbf{87.94$\pm$0.13} \\
        MaC-Cal
        & \underline{0.60$\pm$0.03} & \underline{0.81$\pm$0.01} & 97.80$\pm$0.01
        & \underline{3.85$\pm$0.26} & \underline{3.77$\pm$0.34} & 87.03$\pm$0.13  \\
        \textbf{EUA-Cal}
        & \textbf{0.48$\pm$0.08} & \textbf{0.20$\pm$0.03} & 97.43$\pm$0.01
        & \textbf{0.89$\pm$0.07} & \textbf{0.84$\pm$0.13} & 87.54$\pm$0.12  \\
        \bottomrule
    \end{NiceTabular}%
    }
\end{table*}

\begin{figure}[t]
    \centering

    \subfloat[CIFAR-10]{%
        \includegraphics[width=0.25\linewidth]{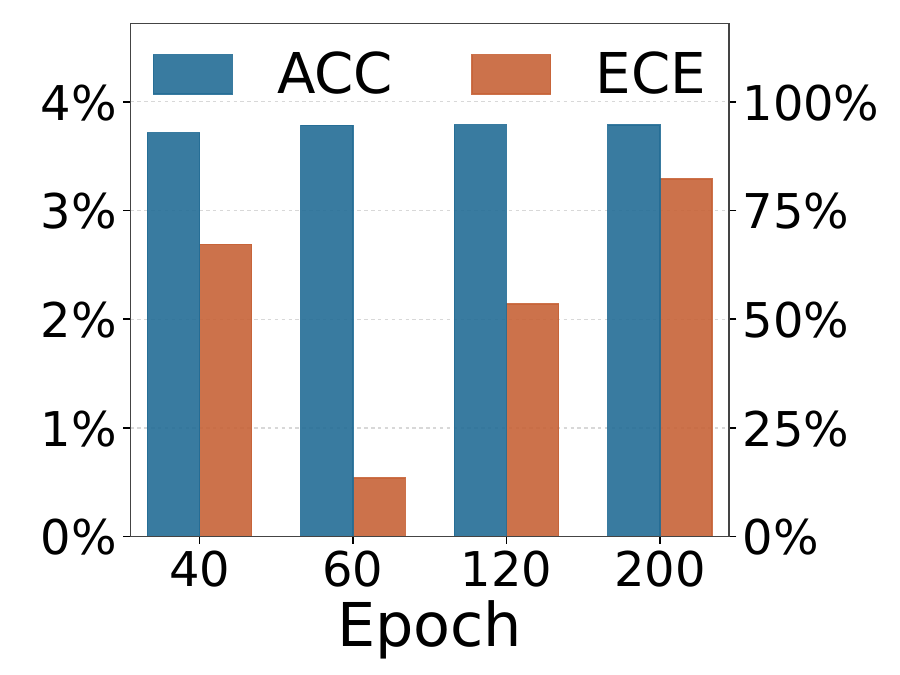}
        \label{7a}
    }
    \subfloat[CIFAR-100]{%
        \includegraphics[width=0.25\linewidth]{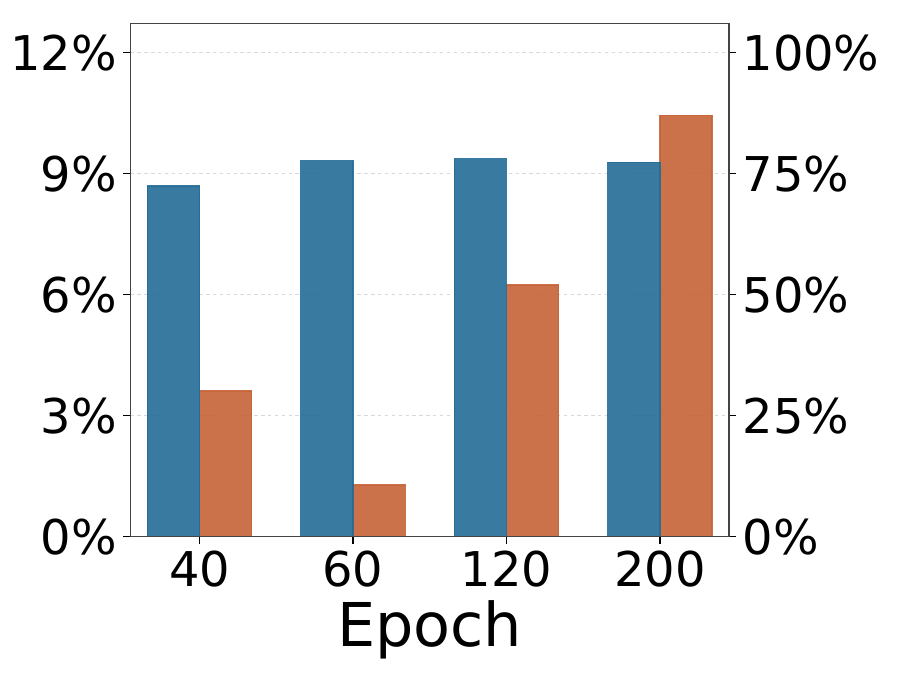}
        \label{7b}
    }
    \subfloat[ARC-C]{%
        \includegraphics[width=0.25\linewidth]{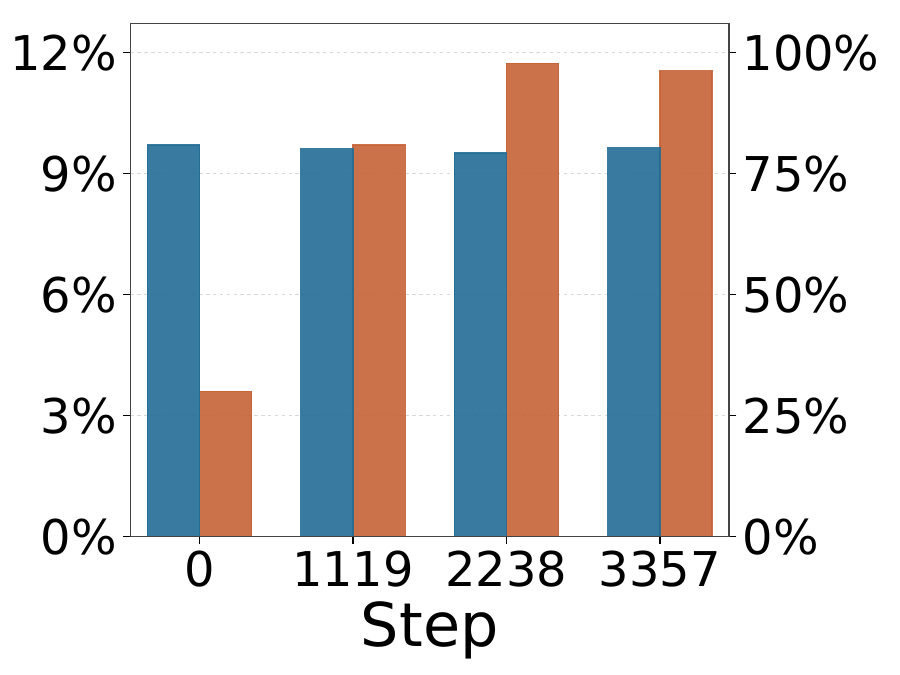}
        \label{7c}
    }
    \subfloat[ARC-E]{%
        \includegraphics[width=0.25\linewidth]{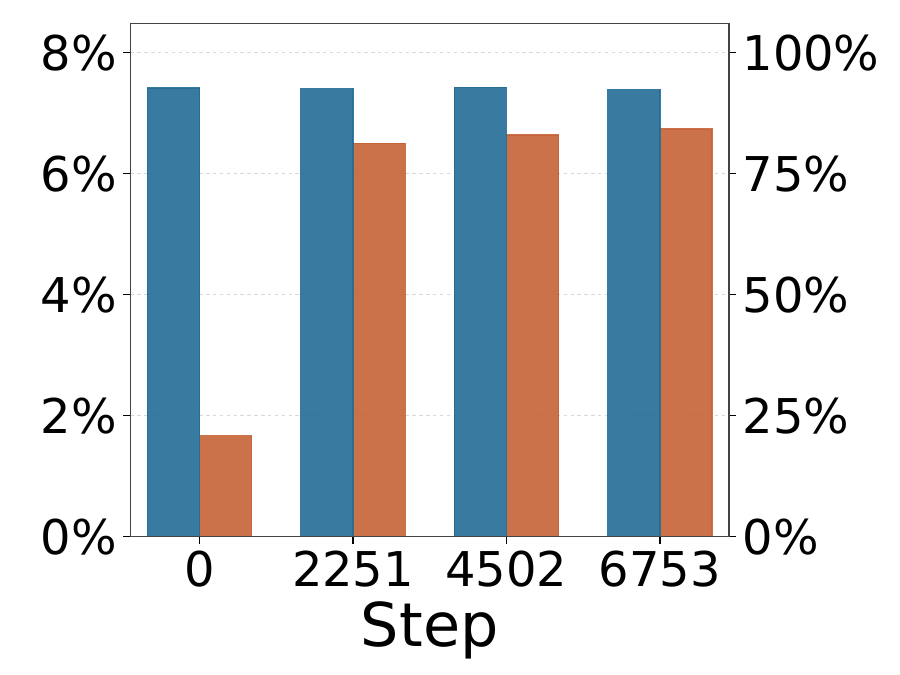}
        \label{7d}
    }

    \caption{Impact of anchor model selection. The horizontal axis indicates the epoch at which the anchor model is selected for ResNet-50 on CIFAR-10 and CIFAR-100, or the optimization step for Llama3-8B on ARC-C and ARC-E.}
    \label{fig7}
\end{figure}

\begin{figure}[!t]
    \centering

    \subfloat[ResNet-50 on CIFAR-100]{%
        \includegraphics[width=0.51\linewidth]{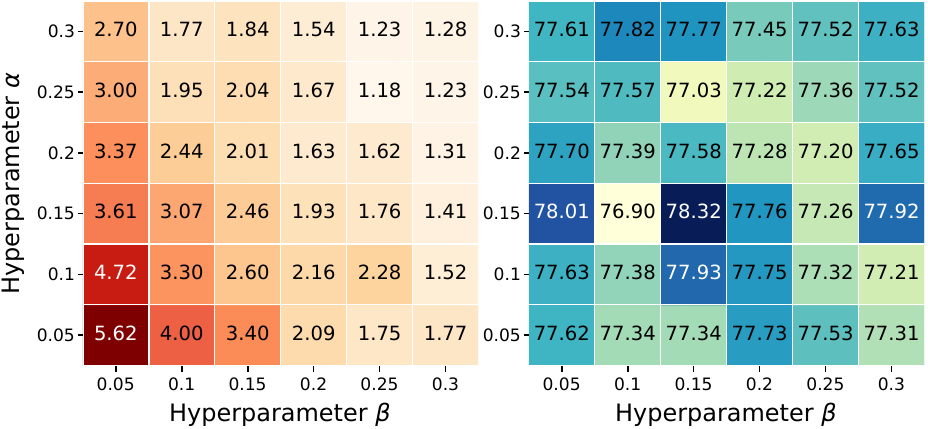}
        \label{8a}
    }
    \subfloat[Llama3-8B on ARC-C]{%
        \includegraphics[width=0.49\linewidth]{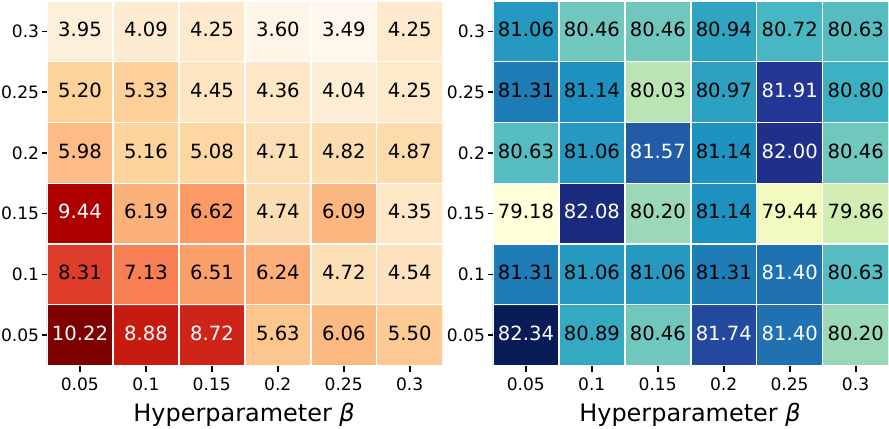}
        \label{8b}
    }
    \caption{Sensitivity to $\alpha$ and $\beta$ of EUA-Cal. ECE ($\downarrow$) is shown in \textcolor{orange}{warmer colors}, and ACC ($\uparrow$) is shown in \textcolor{blue}{cooler colors}.}
    \label{fig8}
\end{figure}

\begin{table*}[t]
    \centering
    \caption{Performance comparison between different methods with Llama3.1-8B on ARC-C, ARC-E, and SciQ datasets. Results are reported as mean $\pm$ standard deviation (\%). The best and second best results are shown in \textbf{bold} and \underline{underlined}, respectively.}
    \label{tb10}
    \setlength{\tabcolsep}{4.0pt}
    \renewcommand{\arraystretch}{1.08}
    \resizebox{\textwidth}{!}{%
    \begin{NiceTabular}{l|ccc|ccc|ccc|c}
        \CodeBefore
        \rowcolor{oursbg}{14}
        \Body
        \toprule
        \multirow{2}{*}{\textbf{Method}}
        & \multicolumn{3}{c}{\textbf{ARC-C}}
        & \multicolumn{3}{c}{\textbf{ARC-E}}
        & \multicolumn{3}{c}{\textbf{SciQ}}
        & \multirow{2}{*}{\shortstack{\textbf{Avg.}\\\textbf{Rank}}} \\
        & ECE $\downarrow$ & AECE $\downarrow$ & ACC $\uparrow$
        & ECE $\downarrow$ & AECE $\downarrow$ & ACC $\uparrow$
        & ECE $\downarrow$ & AECE $\downarrow$ & ACC $\uparrow$ & \\
        \midrule
        Vanilla
        & 17.36$\pm$0.59 & 17.29$\pm$0.53 & 81.54$\pm$0.38
        & 6.75$\pm$0.11 & 6.73$\pm$0.07 & 92.76$\pm$0.11
        & 5.83$\pm$0.10 & 5.72$\pm$0.13 & 94.03$\pm$0.12 & 10.00 \\
        LS
        & 14.30$\pm$0.27 & 14.67$\pm$0.35 & 80.66$\pm$0.34
        & \textbf{3.01$\pm$0.08} & 5.11$\pm$0.23 & 92.55$\pm$0.11
        & \textbf{1.79$\pm$0.32} & 4.57$\pm$0.29 & 94.43$\pm$0.31 & 5.56 \\
        FL
        & 12.40$\pm$2.18 & 12.74$\pm$1.64 & 80.97$\pm$0.12
        & 4.05$\pm$0.25 & \underline{4.03$\pm$0.08} & \underline{92.92$\pm$0.09}
        & 3.50$\pm$0.17 & 4.21$\pm$0.17 & 94.27$\pm$0.31 & \underline{4.56} \\
        CRL
        & 16.26$\pm$1.29 & 16.01$\pm$1.38 & 82.30$\pm$0.18
        & 5.80$\pm$0.09 & 5.71$\pm$0.03 & \textbf{93.10$\pm$0.11}
        & 5.23$\pm$0.15 & 4.94$\pm$0.22 & 94.37$\pm$0.15 & 6.67 \\
        Distillation
        & 9.12$\pm$0.90 & 10.41$\pm$0.58 & 81.48$\pm$0.64
        & \underline{3.03$\pm$0.10} & 4.06$\pm$0.14 & 92.80$\pm$0.15
        & 3.81$\pm$0.18 & 4.97$\pm$0.22 & 94.10$\pm$0.10 & 5.00 \\
        MbLS
        & 18.41$\pm$1.20 & 18.33$\pm$1.17 & 80.69$\pm$1.16
        & 6.73$\pm$0.03 & 6.64$\pm$0.08 & 92.87$\pm$0.09
        & 5.63$\pm$0.28 & 5.51$\pm$0.33 & 94.23$\pm$0.29 & 9.56 \\
        CSAM
        & 15.25$\pm$0.42 & 15.24$\pm$0.42 & \textbf{83.22$\pm$0.43}
        & 4.85$\pm$0.21 & 4.77$\pm$0.15 & 92.42$\pm$0.11
        & 3.87$\pm$0.10 & \underline{3.73$\pm$0.08} & 94.53$\pm$0.15 & 5.56 \\
        Mixup
        & 12.76$\pm$0.57 & 12.96$\pm$0.57 & 80.20$\pm$0.30
        & 6.07$\pm$0.28 & 6.07$\pm$0.26 & 92.17$\pm$0.15
        & 4.52$\pm$0.38 & 4.54$\pm$0.40 & \textbf{95.00$\pm$0.46} & 7.78 \\
        MIT-A
        & 7.87$\pm$0.45 & \underline{7.79$\pm$0.45} & 80.57$\pm$0.30
        & 5.80$\pm$0.22 & 5.73$\pm$0.15 & 92.66$\pm$0.28
        & 4.55$\pm$0.30 & 4.40$\pm$0.42 & \underline{94.77$\pm$0.31} & 5.78 \\
        EDL
        & 11.15$\pm$0.19 & 12.07$\pm$0.63 & 80.69$\pm$0.40
        & 4.26$\pm$0.53 & 5.29$\pm$0.48 & 92.45$\pm$0.25
        & 1.88$\pm$0.36 & 5.42$\pm$0.46 & 94.37$\pm$0.38 & 6.00 \\
        IB-EDL
        & \underline{4.86$\pm$0.29} & 9.82$\pm$0.32 & 80.77$\pm$0.21
        & 6.30$\pm$0.26 & 6.34$\pm$0.20 & 92.35$\pm$0.06
        & 1.97$\pm$0.18 & 5.82$\pm$0.20 & 94.10$\pm$0.10 & 7.44 \\
        \textbf{EUA-Cal}
        & \textbf{3.82$\pm$0.49} & \textbf{3.95$\pm$0.30} & \underline{82.42$\pm$0.39}
        & 3.38$\pm$0.05 & \textbf{3.34$\pm$0.19} & 92.35$\pm$0.11
        & \underline{1.86$\pm$0.14} & \textbf{1.83$\pm$0.25} & 94.10$\pm$0.17 & \textbf{3.33} \\
        \bottomrule
    \end{NiceTabular}%
    }
\end{table*}

\begin{table*}[t]
    \centering
    \caption{Performance comparison between different methods with Qwen3-8B-Base on ARC-C, ARC-E, and SciQ datasets. Results are reported as mean $\pm$ standard deviation (\%). The best and second best results are shown in \textbf{bold} and \underline{underlined}, respectively.}
    \label{tb11}
    \setlength{\tabcolsep}{4.0pt}
    \renewcommand{\arraystretch}{1.08}
    \resizebox{\textwidth}{!}{%
    \begin{NiceTabular}{l|ccc|ccc|ccc|c}
        \CodeBefore
        \rowcolor{oursbg}{14}
        \Body
        \toprule
        \multirow{2}{*}{\textbf{Method}}
        & \multicolumn{3}{c}{\textbf{ARC-C}}
        & \multicolumn{3}{c}{\textbf{ARC-E}}
        & \multicolumn{3}{c}{\textbf{SciQ}}
        & \multirow{2}{*}{\shortstack{\textbf{Avg.}\\\textbf{Rank}}} \\
        & ECE $\downarrow$ & AECE $\downarrow$ & ACC $\uparrow$
        & ECE $\downarrow$ & AECE $\downarrow$ & ACC $\uparrow$
        & ECE $\downarrow$ & AECE $\downarrow$ & ACC $\uparrow$ & \\
        \midrule
        Vanilla
        & 8.10$\pm$0.71 & 7.89$\pm$0.70 & 91.50$\pm$0.56
        & 2.41$\pm$0.02 & 2.32$\pm$0.07 & 97.31$\pm$0.08
        & 3.64$\pm$0.15 & 3.51$\pm$0.15 & 96.27$\pm$0.12 & 10.22 \\
        LS
        & 3.53$\pm$0.03 & 5.62$\pm$0.25 & 91.75$\pm$0.27
        & 3.86$\pm$0.18 & 3.94$\pm$0.10 & 97.40$\pm$0.10
        & 2.75$\pm$0.39 & 4.42$\pm$0.09 & 96.53$\pm$0.23 & 8.22 \\
        FL
        & \underline{3.33$\pm$0.23} & 3.60$\pm$0.25 & 91.70$\pm$0.05
        & \textbf{1.21$\pm$0.14} & \textbf{1.01$\pm$0.04} & 97.45$\pm$0.07
        & \underline{1.67$\pm$0.39} & 2.31$\pm$0.44 & 96.20$\pm$0.40 & 4.44 \\
        CRL
        & 5.00$\pm$0.11 & 5.14$\pm$0.06 & 91.70$\pm$0.05
        & 1.36$\pm$0.08 & 1.37$\pm$0.10 & 97.46$\pm$0.09
        & 3.07$\pm$0.13 & 2.87$\pm$0.42 & 96.10$\pm$0.36 & 7.33 \\
        Distillation
        & 3.77$\pm$0.11 & 4.57$\pm$0.26 & 91.87$\pm$0.20
        & 1.47$\pm$0.03 & 1.46$\pm$0.02 & 96.97$\pm$0.00
        & 2.60$\pm$0.31 & 3.74$\pm$0.19 & 96.07$\pm$0.06 & 7.44 \\
        MbLS
        & 6.93$\pm$0.19 & 6.67$\pm$0.08 & 92.18$\pm$0.13
        & 2.25$\pm$0.04 & 2.25$\pm$0.08 & 97.38$\pm$0.09
        & 3.28$\pm$0.07 & 2.96$\pm$0.08 & 96.77$\pm$0.12 & 7.78 \\
        CSAM
        & 4.02$\pm$0.20 & 3.62$\pm$0.08 & 92.28$\pm$0.06
        & \underline{1.27$\pm$0.07} & 1.15$\pm$0.05 & \underline{97.66$\pm$0.05}
        & 2.01$\pm$0.22 & \underline{1.52$\pm$0.09} & \textbf{97.13$\pm$0.06} & \underline{3.00} \\
        Mixup
        & 4.41$\pm$0.23 & 4.93$\pm$0.18 & 92.09$\pm$0.10
        & 1.33$\pm$0.15 & 1.76$\pm$0.13 & 97.53$\pm$0.13
        & 2.91$\pm$0.03 & 2.89$\pm$0.03 & 96.63$\pm$0.06 & 5.89 \\
        MIT-A
        & 3.71$\pm$0.64 & \underline{3.48$\pm$0.28} & \textbf{92.89$\pm$0.21}
        & 1.56$\pm$0.11 & 1.44$\pm$0.05 & \textbf{97.97$\pm$0.02}
        & 2.61$\pm$0.14 & 2.30$\pm$0.23 & \underline{97.07$\pm$0.21} & 3.44 \\
        EDL
        & 5.48$\pm$0.30 & 5.91$\pm$0.63 & \underline{92.35$\pm$0.32}
        & 3.02$\pm$0.19 & 3.69$\pm$0.32 & 97.35$\pm$0.18
        & 1.80$\pm$0.37 & 3.97$\pm$0.10 & 96.67$\pm$0.06 & 7.56 \\
        IB-EDL
        & 8.32$\pm$0.14 & 10.97$\pm$0.09 & 92.18$\pm$0.05
        & 3.87$\pm$0.24 & 4.67$\pm$0.16 & 97.57$\pm$0.16
        & 2.96$\pm$0.67 & 4.80$\pm$0.16 & 96.43$\pm$0.25 & 9.33 \\
        \textbf{EUA-Cal}
        & \textbf{2.04$\pm$0.46} & \textbf{2.68$\pm$0.10} & 91.72$\pm$0.26
        & 1.33$\pm$0.08 & \underline{1.11$\pm$0.06} & 97.49$\pm$0.13
        & \textbf{1.55$\pm$0.42} & \textbf{1.25$\pm$0.12} & 96.77$\pm$0.15 & \textbf{2.89} \\
        \bottomrule
    \end{NiceTabular}%
    }
\end{table*}

\subsection{Impact of Anchor Model Selection}
\label{C2}
Figure~\ref{fig7} illustrates how the selection of anchor affects EUA-Cal. On CIFAR-10 and CIFAR-100, the model at epoch 60 yields the lowest ECE. Selecting the model at epoch 40 results in lower accuracy, whereas later models substantially increase the ECE with marginal accuracy gain. For ARC-C and ARC-E, using the pretrained Llama3-8B as the anchor model yields the lowest ECE. Later models increase the ECE, while the accuracy changes only slightly. These results indicate that the selection of the anchor model should consider the balance between classification performance and calibration performance, rather than simply selecting the earliest model.

\subsection{Impact of Hyperparameters}
\label{C3}
\textbf{Sensitivity to $\alpha$ and $\beta$.} Figure~\ref{fig8} illustrates that our EUA-Cal method improves calibration performance for most settings of $\alpha$ and $\beta$ on both CIFAR-100 and ARC-C, while ACC remains relatively stable. This indicates that the calibration improvement of our method is not restricted to a particular choice of $\alpha$ and $\beta$, suggesting that EUA-Cal is robust to these hyperparameters.

\begin{figure}[t]
    \centering

    \subfloat[ResNet-50 on CIFAR-10]{%
        \includegraphics[width=0.5\linewidth]{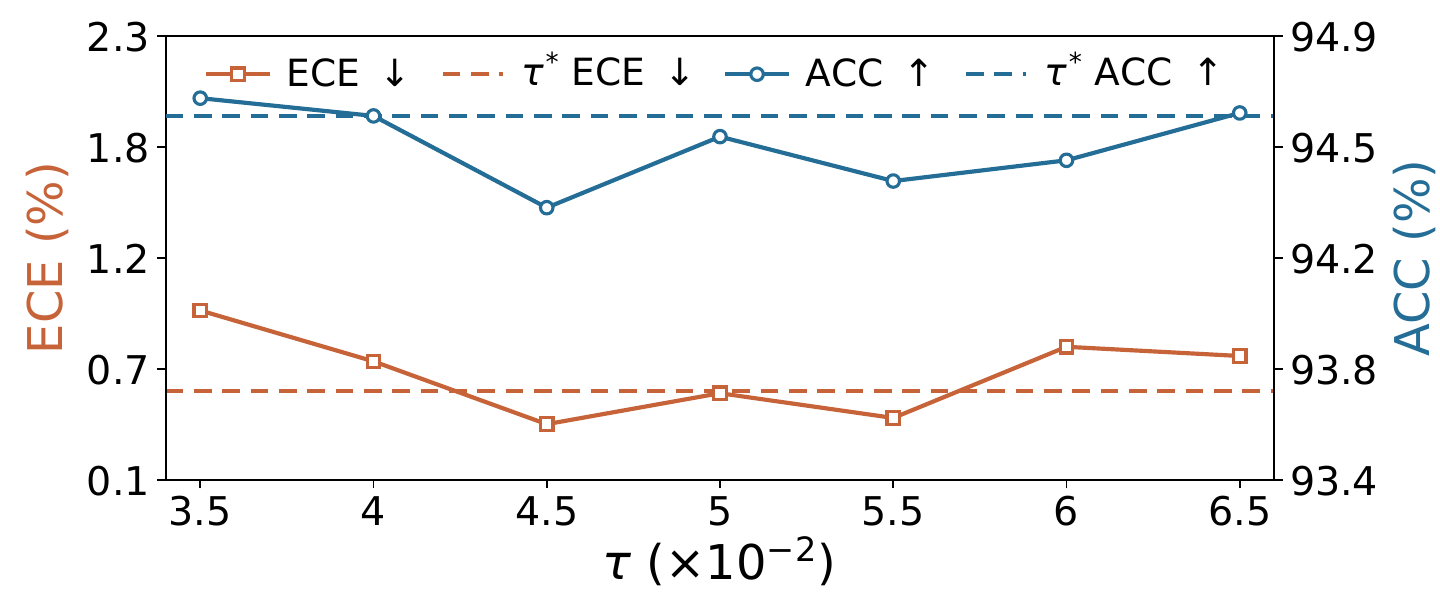}
        \label{9a}
    }
    \subfloat[ResNet-50 on CIFAR-100]{%
        \includegraphics[width=0.5\linewidth]{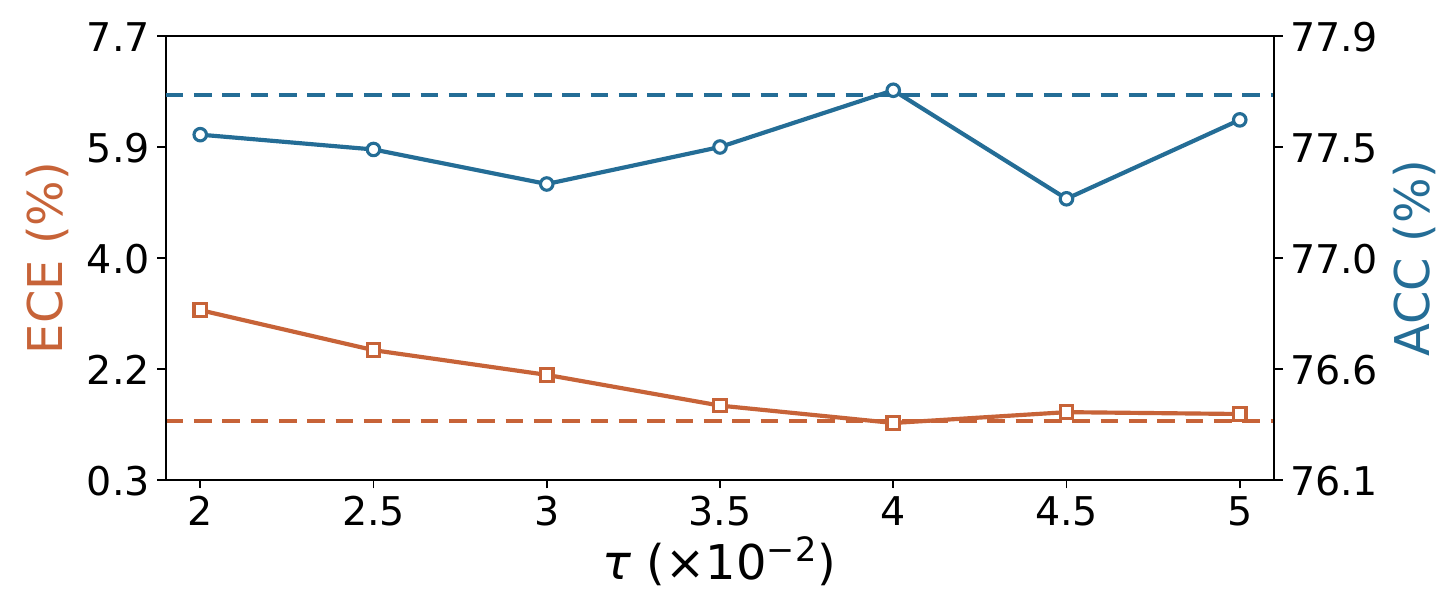}
        \label{9b}
    }\par
    \subfloat[Llama3-8B on ARC-C]{%
        \includegraphics[width=0.5\linewidth]{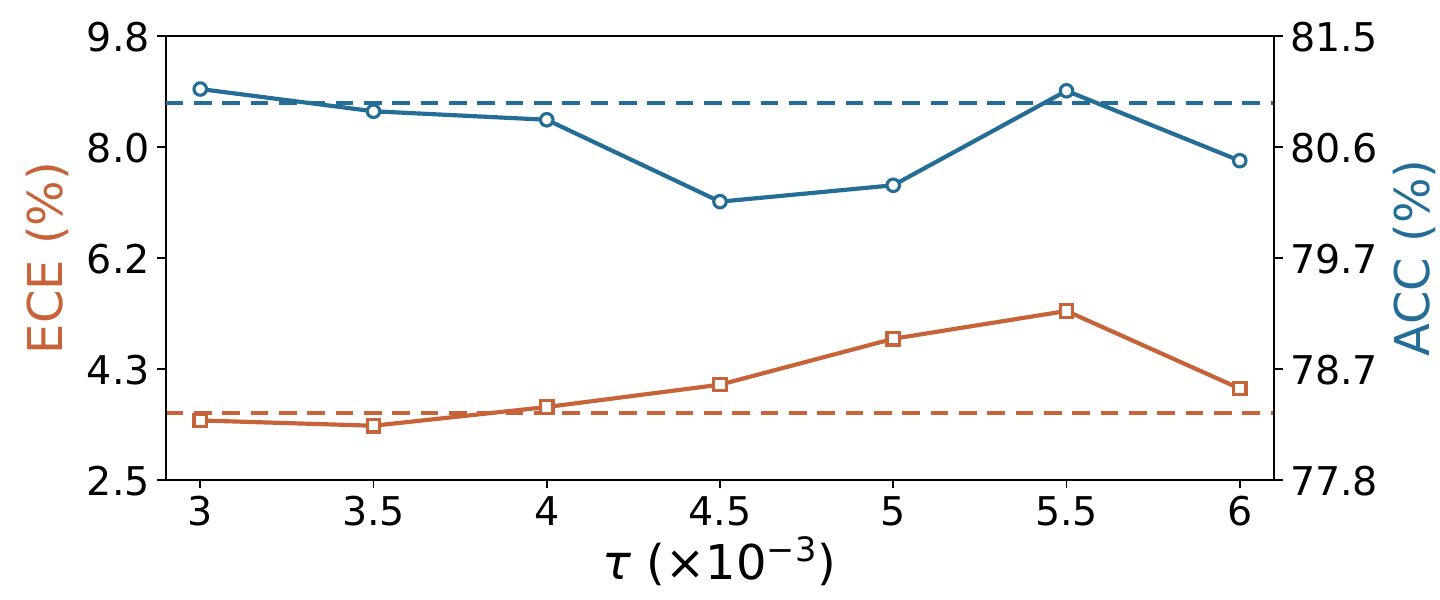}
        \label{9c}
    }
    \subfloat[Llama3-8B on ARC-E]{%
        \includegraphics[width=0.5\linewidth]{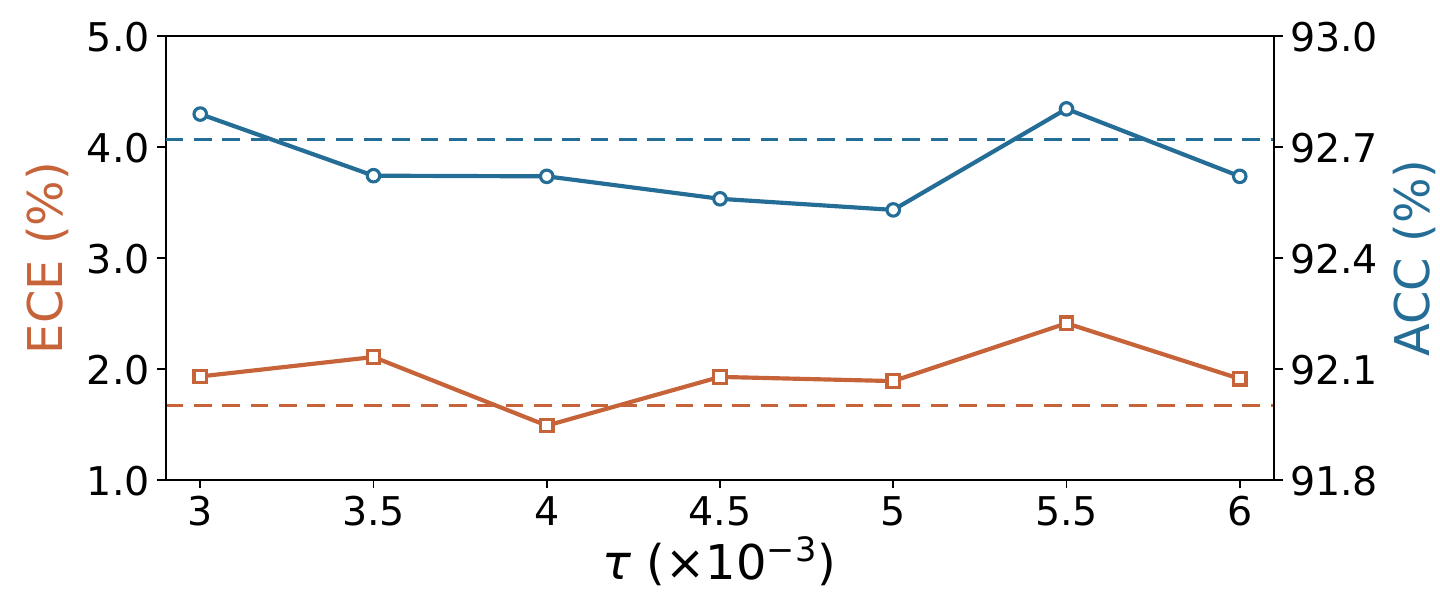}
        \label{9d}
    }

    \caption{Sensitivity to $\tau$ of EUA-Cal. ECE ($\downarrow$) is shown in \textcolor{orange}{warmer colors}, and ACC ($\uparrow$) is shown in \textcolor{blue}{cooler colors}.}
    \label{fig9}
\end{figure}

\textbf{Sensitivity to $\tau$.} Figure~\ref{fig9} reports the ACC and ECE of our EUA-Cal method under different values of $\tau$ on CIFAR-10, CIFAR-100, ARC-C, and ARC-E. As $\tau$ varies, ACC remains relatively stable, while ECE varies more. Despite this variation, selecting $\tau^*$ by minimizing training-set NLL achieves competitive ECE on all four datasets without a separate search over $\tau$.

\end{document}